\documentclass{article}

\usepackage[utf8]{inputenc}
\usepackage{combelow}
\usepackage{hyperref}
\usepackage{graphicx}
\usepackage{subcaption}
\usepackage[para]{threeparttable}
\usepackage{cite}
\usepackage{float}
\usepackage{dblfloatfix}
\usepackage{booktabs}
\usepackage{amsthm}
\usepackage{microtype}
\usepackage{authblk}
\usepackage{fullpage}

\newtheorem{theorem}{Theorem}
\newtheorem{lemma}{Lemma}

\title{Temporal Coding in Spiking Neural Networks with Alpha Synaptic Function: Learning with Backpropagation}
\author[1]{Iulia~M.~Com\cb{s}a\footnote{Correspondence: iuliacomsa@google.com}}
\author[1]{Krzysztof~Potempa}
\author[1]{Luca~Versari}
\author[1]{Thomas~Fischbacher}
\author[1]{Andrea~Gesmundo}
\author[1]{Jyrki~Alakuijala}
\affil[1]{Google Research Z{\"u}rich, Switzerland}

\begin{document}

\maketitle


\begin{abstract}
The timing of individual neuronal spikes is essential for biological brains to
  make fast responses to sensory stimuli. However, conventional artificial
  neural networks lack the intrinsic temporal coding ability present in
  biological networks. We propose a spiking neural network model that encodes
  information in the relative timing of individual spikes. In classification
  tasks, the output of the network is indicated by the first neuron to spike in
  the output layer. This temporal coding scheme allows the supervised training
  of the network with backpropagation, using locally exact derivatives of the
  postsynaptic spike times with respect to presynaptic spike times. The network
  operates using a biologically-plausible alpha synaptic transfer function.
  Additionally, we use trainable synchronisation pulses that provide bias, add
  flexibility during training and exploit the decay part of the alpha function.
  We show that such networks can be successfully trained on noisy Boolean logic
  tasks and on the MNIST dataset encoded in time. We show that the spiking
  neural network outperforms comparable spiking models on MNIST and achieves
  similar quality to fully connected conventional networks with the same
  architecture. The spiking network spontaneously discovers two operating
  modes, mirroring the accuracy-speed trade-off observed in human
  decision-making: a highly accurate but slow regime, and a fast but slightly
  lower-accuracy regime. These results demonstrate the computational power of
  spiking networks with biological characteristics that encode information in
  the timing of individual neurons. By studying temporal coding in spiking
  networks, we aim to create building blocks towards energy-efficient,
  state-based and more complex biologically-inspired neural architectures.
\end{abstract}

\section{Introduction}

Inspired by the biology of the nervous system, artificial neural networks have
recently been used to achieve resounding success in solving many real-world
problems, sometimes surpassing the performance of humans~\cite{Silver2016,
Krizhevsky2012, Szegedy2015}. However, conventional artificial networks lack
the intrinsic ability to encode information using temporal coding schemes in
the same way as biological brains do. In the nervous system, the timing of
individual neuronal spikes is fundamental during behaviours requiring rapid
encoding and processing of perceptual stimuli. The relative firing time of
neurons has been shown to encode stimulus information in the
visual~\cite{Gollisch2008}, auditory~\cite{Moiseff1981} and
tactile~\cite{Johansson2004} cortices, but also in higher-level neural
structures like the thalamus~\cite{Reinagel2000} and
hippocampus~\cite{Huxter2003}. Moreover, by observing in-vivo the low latency
of responses to visual stimuli in the temporal cortex, it can be concluded that
the response is produced by individual spikes occurring at every synaptic stage
across the visual areas of the brain~\cite{Thorpe1989}.

Architectures for atemporal networks that emulate the processing of information
in a temporal fashion using memory mechanisms have been
proposed~\cite{Hochreiter1997}. However, these do not have the advantages
conferred by encoding information in temporal domain. One disadvantage is that
every neuron needs to wait for the activation of all the neurons in the
previous layer in order to produce an answer. Compared to biological brains,
this is inefficient in terms of energy consumption. Moreover, information in
the real world almost always has a temporal dimension. Therefore, additional
processing is needed for encoding it in an atemporal network, potentially
losing information in the process.

Unlike the field of conventional artificial neural networks, the
field of artificial spiking networks that encode information in time has not
been thoroughly explored. So far, spiking networks have only achieved modest
results. The main difficulty in advancing the field of spiking networks has
been their training process, as the techniques usually used for supervised
learning in atemporal networks cannot be directly applied when information is
encoded in a temporal sequence of asynchronous events. Atemporal networks owe
much of their success to the development of the backpropagation
algorithm~\cite{Speelpenning1980, Rummelhart1987}. Backpropagation exploits the
existence of an end-to-end differentiable relationship between a loss function
and the network inputs and outputs, which can be expressed in terms of local
derivatives at all hidden network parameters. One can thus find local updates
that minimise the loss function and apply them as incremental updates to train
the network. On the other hand, in spiking networks, which encode information
in sequences of binary spike events, differentiable relationships do not
naturally exist.

The problem of training in spiking networks has been addressed in several ways.
Many spiking models have adopted a rate coding scheme. In contrast with
temporal coding of information, which is based on individual spike timing, rate
coding averages over multiple spikes. Approximate gradients have been proposed
in order to allow gradient descent optimization in such networks~\cite{Wu2018,
Bellec2018, Lee2019}. Various other learning rules have been proposed with the
objective of producing custom spiking patterns or spike
distributions~\cite{Sussillo2009, Xu2013, Florian2012, Gutig2006, Huh2017,
Zenke2018, Neftci2017, Sporea2013, Esser2015, Sengupta2019}. Alternatively,
atemporal deep neural networks can be trained and then converted to spiking
networks using rate-coding schemes~\cite{Hunsberger2016, Rueckauer2018,
Diehl2016, Hu2018}. Spiking networks can also be trained using methods such as
evolutionary algorithms~\cite{Pavlidis2005} or reinforcement
learning~\cite{Florian2007}. However, such rate-coding schemes may still be
redundant considering the evidence that biological systems can react to stimuli
on the basis of single spikes~\cite{Thorpe1989, Gollisch2008, Johansson2004}.

In contrast with rate-coding models, here we are interested in the temporal
encoding of information into single spikes. Crucially, this change in coding
scheme shifts the differentiable relationships into the temporal
domain. To find a backpropagation scheme for temporal coding, we need a
differentiable relationship of the time of a postsynaptic spike with respect to
the weights and times of the presynaptic spikes. Encoding information in
temporal domain makes this possible.

A similar idea was proposed in the SpikeProp
model~\cite{Bohte2000}. In this model, a spiking network successfully learns the
supervised XOR problem encoded in temporal domain using neurons that generate
single spikes. The model was able to implement backpropagation by approximating
the differentiable relationship for small learning rates. It was also necessary
to explicitly encode inhibitory and excitatory neurons in order for training to
converge. Various extensions to SpikeProp have been proposed~\cite{Hong2017,
Schrauwen2004, McKennoch2006, Booij2005}. Recently, Mostafa~\cite{Mostafa2017}
trained a spiking network to solve temporally encoded XOR and MNIST problems by
deriving locally exact gradients with non-leaky spiking neurons. Other training
approaches include spike-timing-dependent plasticity~\cite{Kheradpisheh2018} and
reinforcement learning~\cite{Mozafari2018}.

One important choice when modelling a spiking neural network is that of the
synaptic transfer function, which defines the dynamics of the neuron membrane
potential in response to a presynaptic spike. From a machine learning
perspective, this is equivalent to the activation function in conventional
networks, but, importantly, it operates in time. Historically, the
Hodgkin-Huxley model~\cite{Hodgkin1952} was the first to offer a detailed
description of the process of neuronal spiking using differential equations
describing the dynamics of sodium and potassium ionic channels. In practice,
however, this model is often needlessly complex. Instead, many applications use
the reduced Spike Response Model (SRM)~\cite{Gerstner2001}. In this model, the
membrane potential is described by the integration of kernels reflecting the
incoming currents at the synapse. The neuron spikes when its membrane potential
reaches a threshold, then enters a refractory period. Provided that an
appropriate kernel function is used~\cite{Sterratt2018}, the SRM can
approximate the dynamics of the Hodgkin-Huxley model to a satisfactory extent,
while demanding less computational power and being easier to analyse.
Integrate-and-fire neurons and their leaky counterparts are examples of SRM. A
commonly used SRM impulse response function is the exponential decay, $e^{-t}$.

A more biologically-realistic but less explored alternative is the alpha
function, of the form $t e^{-t}$, produced by the integration of exponentially
decaying kernels. In contrast with the single exponential function, the alpha
function gradually rises before slowly decaying (Figure~\ref{fig:1A}), which
allows more intricate interactions between presynaptic inputs. The alpha
function has been proposed~\cite{Rall1967} as a close match to the shape of
postsynaptic potentials measured in vitro~\cite{Frank1959, Burke1967}. It hence
provides a biologically-plausible model for exploring the problem-solving
abilities of spiking networks with temporal coding schemes. As we show below,
it is possible to derive exact gradients with respect to spike times using this
model.

With these considerations in mind, here we propose a spiking network model that
uses the alpha function for synaptic transfer and encodes information in
relative spike times. The network is fully trained in temporal domain using
exact gradients over domains where relative spiking order is preserved. To our
knowledge, this model has not been previously used for supervised learning with
backpropagation. We explore the capacity of this model to learn
standard benchmark problems, such as Boolean logic gates and MNIST, encoded in
individual spike times. To facilitate transformations of the class boundaries,
we use synchronisation pulses, which are neurons that send spikes at
input-independent, learned times. The model is easily able to solve
temporally-encoded Boolean logic and other benchmark problems. We perform a
search for the best set of hyperparameters for this model using
evolutionary-neural hybrid agents~\cite{Maziarz2018} in order to solve
temporally-encoded MNIST. The result improves the state-of-the-art accuracy on
non-convolutional spiking networks and is comparable to the performance of
atemporal non-convolutional networks. Furthermore, we analyse the behaviour of
the spiking network during training and show that it spontaneously displays two
operational regimes that reflect a trade-off between speed and accuracy: a slow
regime that is slow but very accurate, and a fast regime that is slightly less
accurate but makes decisions much faster.

With this work, we also aim to increase the familiarity of the research
community with the concept of temporal coding in spiking neural networks. Given
that biological brains have evolved for millions of years to use temporal
coding mechanisms in order to process information efficiently, we expect that
an equivalent development will also be a key step in advancing artificial
intelligence in the future. The present work is an early building block in
this direction that invites further exploration of more complex recurrent
architectures, spike-based state machines and interfacing between artificial
and biological spiking neural networks.

\section{Methods}

\subsection{Temporal coding}

In this model, information is encoded in the relative timing of individual
spikes. The input features are encoded in temporal domain as the spike times of
individual input neurons, with each neuron corresponding to a distinct feature.
More salient information about a feature is encoded as an earlier spike in the
corresponding neuron. Information propagates through the network in a temporal
fashion. Each hidden and output neuron spikes when its membrane potential rises
above a fixed threshold. Similarly to the input layer, the output layer of the
network encodes a result in the relative timing of output spikes. In other
words, the computational process consists of producing a temporal sequence of
spikes across the network in a particular order, with the result encoded in the
ordering of spikes in the output layer.

We use this model to solve standard classification problems. Given a
classification problem with $m$ inputs and $n$ possible classes, the inputs are
encoded as the spike times of individual neurons in the input layer and the
result is encoded as the index of the neuron that spikes first among the
neurons in the output layer. An example drawn from class $k$ is classified
correctly if and only if the $k^{th}$ output neuron is the first to spike. An
earlier output spike can reflect more confidence of the network in classifying
a particular example, as it implies more synaptic efficiency or a smaller
number of presynaptic spikes. In a biological setting, the winning neuron could
suppress the activity of neighbouring neurons through lateral
inhibition~\cite{Blakemore1970}, while in a machine learning setting the spike
times of the non-winning neurons can be useful in indicating alternatives
predictions of the network. The learning process aims to change the synaptic
weights and thus the spike timings in such a way that the target order of
spikes is produced.

\subsection{Alpha activation function}
\begin{figure}[ht]
    \centering
    \begin{subfigure}[t]{0.48\textwidth}
        \includegraphics[width=\textwidth]{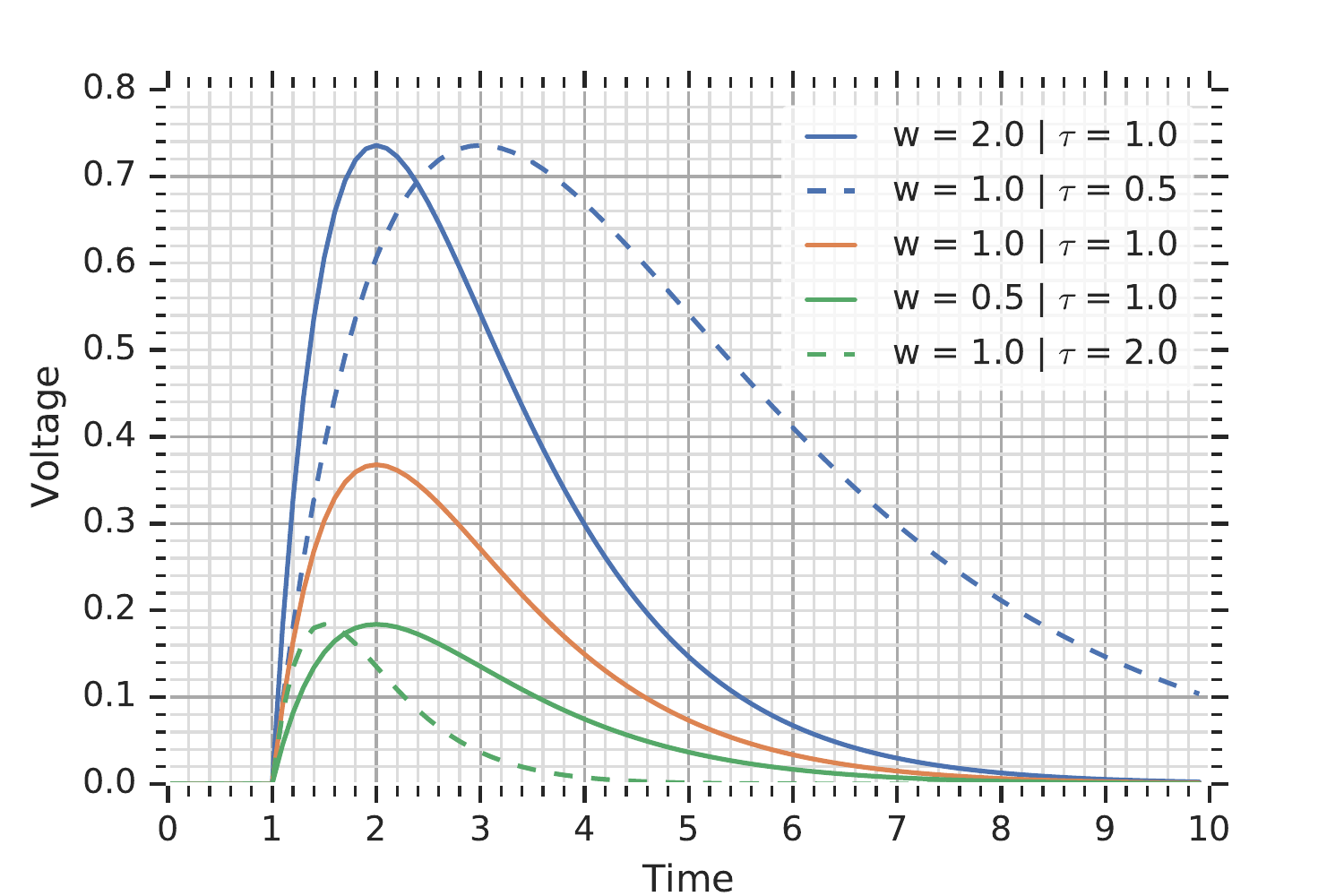}
        \caption{Alpha potential function with different sets of weights $w$
        and decay constants $\tau$. The weight scales the function in
        amplitude, whereas the decay constant scales it in both amplitude and
        time.}
        \label{fig:1A}
    \end{subfigure}
    \quad
    \begin{subfigure}[t]{0.48\textwidth}
        \includegraphics[width=\textwidth]{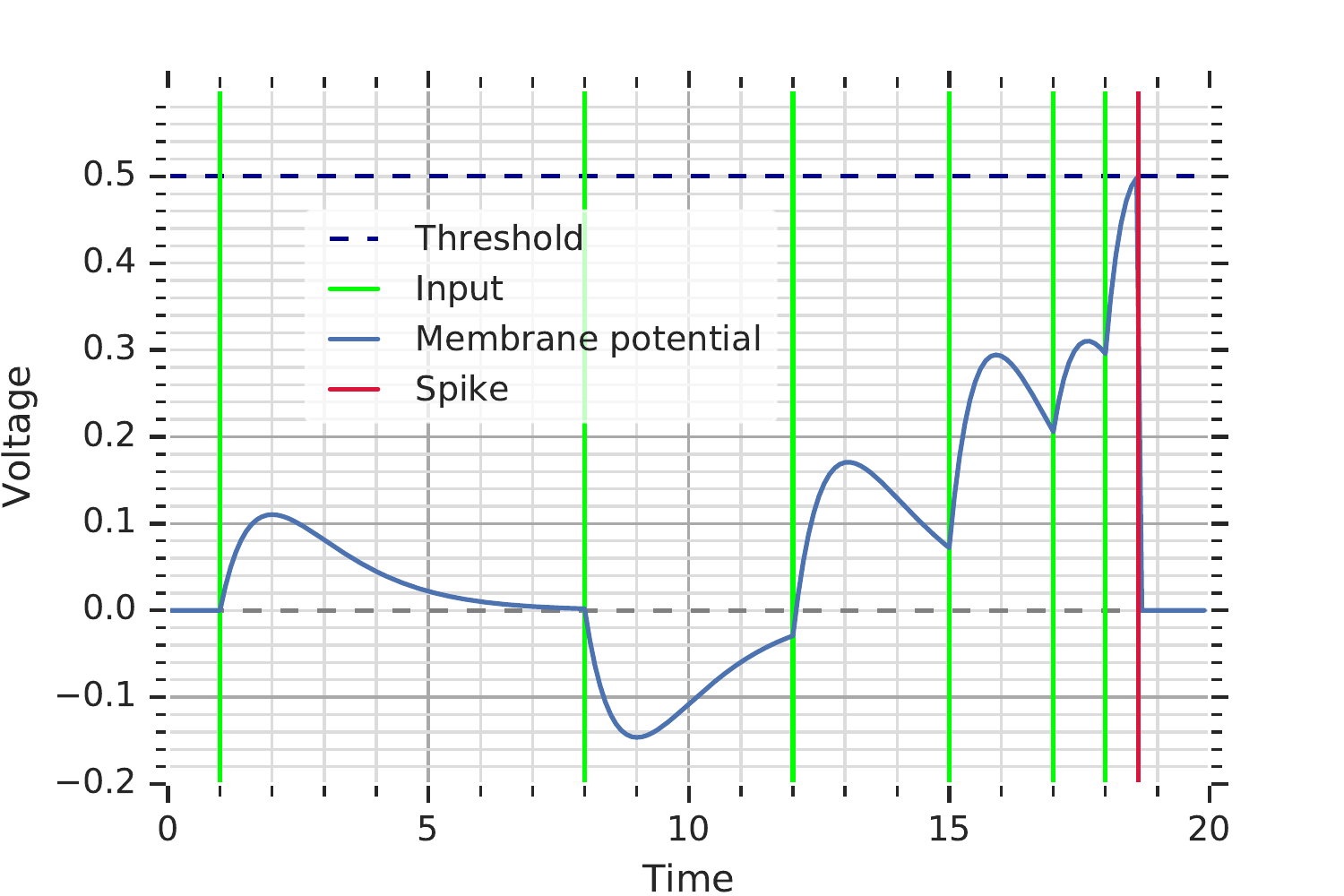}
        \caption{Example of potential membrane dynamics in response to
        excitatory and inhibitory inputs, followed by a spike. In this example,
        $\tau = 1$, $w = \{0.3, -0.4, 0.5, 0.7, 0.5, 0.8\}$, $t = \{1, 8, 12,
        15, 17, 18\}$ and the spike occurs at $t_{out}=18.64$.}
        \label{fig:1B}
    \end{subfigure}
    \caption{\label{fig:1}Description of the neuron model with alpha synaptic function.}
\end{figure}

The neuronal membrane dynamics are governed by a SRM model with alpha function
of synaptic transfer~\cite{Gerstner2001, Sterratt2018, Rall1967}. This is
obtained by integrating over time the incoming exponential synaptic current
kernels of the form $\epsilon(t) = \tau^{-1} e^{-\tau t}$, where $\tau$ is the
decay constant. The potential of the neuronal membrane in response to a single
incoming spike is then of the form $u(t) = t e^{-\tau t}$. This function has a
gradual rise and a slow decay, peaking at $t_{max} = \tau^{-1}$.  Every
synaptic connection has an efficiency, or a weight. The decay rate has the
effect of scaling the induced potential in amplitude and time, while the weight
of the synapse has the effect of scaling the amplitude only
(Figure~\ref{fig:1A}).

Given a set $I$ of presynaptic inputs $i$ arriving at times $t_i \leq t$ with
weights $w_i$ and assuming the postsynaptic neuron has not yet spiked, its
membrane potential at time $t$ is given by:

\begin{equation}
\label{eq:1}
V_{mem}(t)= \sum_{i \in I} w_i (t - t_i) e^{\tau (t_i - t)}
\end{equation}

The neuron spikes when the membrane potential crosses the firing threshold
(Figure~\ref{fig:1B}). To compute the spike time $t_{out}$ of a neuron, we
determine the minimal subset of all presynaptic inputs $I_{t_{out}}$ with $t_i
\leq t_{out}$ which cause the membrane potential to reach the threshold
$\theta$ while rising:

\begin{equation}
\label{eq:2}
\sum_{i \in I_{t_{out}}} w_i (t_{out} - t_i) e^{\tau(t_i - t_{out})} = \theta
\end{equation}

This is achieved by sorting the inputs and adding them to $I_{t_{out}}$ one by
one, until an incoming input arrives later than the predicted spike (if any) or
there are no more inputs. Note that the set $I$ may not simply be computed as
the earliest subset of presynaptic inputs that cause the membrane voltage to
cross $\theta$. If a subset of inputs $I$ causes the membrane to cross $\theta$
at time $t_{out}$, any additional inputs that occur between the maximum
$t_i \in I$ and $t_{out}$ must be considered, and $t_{out}$ must be recomputed.

Eq. \ref{eq:2} has two potential solutions --- one on the rising part of the
function and one on the decaying part. If a solution exists (in other words, if
the neuron spikes), then its spike time is the earlier of the two solutions.

For a set of inputs $I$, we denote $ A_I = \sum_{i \in I} w_i e ^{\tau t_i}$
and $B_I = \sum_{i \in I} w_i e^{\tau t_i} t_i$. The spike time $t_{out}$ can be
computed by solving Eq. \ref{eq:2} using the Lambert $W$
function~\cite{Lambert1758, Corless1996}:

\begin{equation}
\label{eq:3}
t_{out} = \frac{B_I}{A_I} - \frac{1}{\tau} W(-\tau \frac{\theta}{A_I} e ^{\tau \frac{B_I}{A_I}})
\end{equation}

A spike will occur whenever the Lambert $W$ function has a valid argument and
the resulting $t_{out}$ is larger than all input spikes. As we are interested in
the earlier solution of this equation, we employ the main branch of the Lambert
$W$ function. The Lambert $W$ function is real-valued when its argument is
larger than or equal to $-e^{-1}$. It can be proven that this is always the case
when Eq.~\ref{eq:2} has a solution, by expanding $V_{mem}(t_{max}) \geq \theta$,
where $t_{max} = \frac{B}{A} + \frac{1}{\tau}$ is the peak of the membrane
potential function corresponding to the presynaptic set of inputs $I$.

In Appendix~\ref{apx:ua}, we show that the proposed model is powerful enough to
represent any sufficiently well-behaved function.

\subsection{Error backpropagation}

The spiking network learns to solve problems whose inputs and solution are
encoded in the times of individual input and output spikes. Therefore, the goal
is to adjust the output spike times so that their relative order is correct.
Given a classification problem with $n$ classes, the neuron corresponding to
the correct label should be the earliest to spike. We therefore choose a loss
function that minimises the spike time of the target neuron and maximises the
spike time of the non-target neurons. Note that this is the opposite of the
usual classification setting involving probabilities, where the value
corresponding to the correct class is maximised and those corresponding to
incorrect classes are minimised. To achieve this, we use the softmax function
on the \emph{negative} values of the spike times $o_i$ (which are always
positive) in the output layer: $p_j = {e^{-o_j}} / {\sum_{i=1}^{n} e^{-o_i}}$.

We then employ the cross-entropy loss in the usual form: $L(y_i, p_i) =
-\sum_{i=1}^{n}y_i \ln{p_i}$, where $y_i$ is an element of the one-hot encoded
target vector of output spike times. Taking the negative values of the spike
times ensures that minimising the cross-entropy loss minimises the spike time
of the correct label and maximises the rest.

We minimise the cross-entropy loss by changing the value of the weights across
the network. This has the effect of delaying or advancing spike times across
the network. For any presynaptic spike arriving at time $t_j \in I$ with weight
$w_j$, we denote $W_I = W(-\frac{\theta}{A_I} e^\frac{B_I}{A_I})$ and compute
the exact derivative of the postsynaptic spike time with respect to any
presynaptic spike time $t_j$ and its weight $w_j$ as:

\begin{equation}
\label{eq:5}
\frac{\partial t_{out}}{\partial t_j} = \frac{w_j e^{t_j} (t_j - \frac{B_i}{A_I} + W_I + 1)}{A_I (1 + W_I)}
\end{equation}

\begin{equation}
\label{eq:6}
\frac{\partial t_{out}}{\partial w_j} = \frac{e^{t_j} (t_j - \frac{B_I}{A_I} + W_I)}{A_I (1 + W_I)}
\end{equation}

As the postsynaptic spike time moves earlier or later in time, when
$I_{t_{out}}$ changes to include or exclude presynaptic spikes, the landscape of
the loss function also changes. Furthermore, the loss function exhbits
discontinuities where an output neuron stops spiking; we counter this problem
using a penalty, as described below. In practice, we find that optimization is
possible in spite of these challenges.

\subsection{Synchronisation pulses}

In order to adjust the class boundaries in the temporal domain, a temporal form
of bias is also needed to adjust spike times, i.e. to delay or advance them in
time. In this model, we introduce synchronisation pulses acting as additional
inputs across every layer or the network, in order to provide temporal bias
across the network. These can be thought of as similar to internally-generated
rhythmic activity in biological networks, such as alpha waves in the visual
cortex~\cite{Klimesch2012} or theta and gamma waves in the
hippocampus~\cite{Buzsaki2006}.

A set of pulses can be connected to all neurons in the network, to neurons
within individual layers, or to individual neurons. A per-neuron bias is
biologically implausible and more computationally demanding, hence in this
model we use either a single set of pulses per network, to solve easier
problems, or a set of pulses per layer, to solve more difficult problems. All
pulses are fully connected to either all non-input neurons in the network or to
all neurons of the non-input layer they are assigned to.

Each pulse spikes at a predefined and trainable time, providing a reference
spike delay. Each set of pulses is initialised to spike at times evenly
distributed in the interval $(0,1)$. Subsequently, the spike time of each pulse
is learned using Eq. \ref{eq:5}, while the weights between pulses and neurons
are trained using Eq. \ref{eq:6}, in the same way as all other weights in the
network.

\subsection{Hyperparameters}

We trained fully connected feedforward networks with topology
\texttt{n\_hidden} (a vector of hidden layer sizes). We used Adam
optimization~\cite{Kingma2014} with mini-batches of size \texttt{batch\_size}
to minimise the cross-entropy loss. The Adam optimizer performed better than
stochastic gradient descent. We used different learning rates for the pulse
spike time (\texttt{learning\_rate\_pulses}) and the weights of both pulse and
non-pulse neurons (\texttt{learning\_rate}). We used a fixed firing threshold
(\texttt{fire\_threshold}) and decay constant (\texttt{decay\_constant}).

Network weight initialisation is crucial for the subsequent training of the
network. In a spiking network, it is important that the initial weights are
large enough to cause at least some of the neurons to spike; in absence of
spike events, there will be no gradient to use for learning. We therefore used
a modified form of Glorot initialization~\cite{Glorot2010} where the weights
are drawn from a normal distribution with standard deviation $\sigma =
\sqrt{2.0 / (fan_{in} + fan_{out})}$ (as in the original scheme) and custom
mean $\mu = \texttt{multiplier} \times \sigma$. If the multiplication factor of
the mean is $0$, this is the same is the original Glorot initialization scheme.
We set different multiplication factors for pulse
(\texttt{pulse\_init\_multiplier}), and non-pulse
(\texttt{nonpulse\_init\_multiplier}) weights. This allows the two types of
neurons to pre-specialise into inhibitory and excitatory roles. In biological
brains, internal oscillations are thought to be generated through inhibitory
activities that regulate the excitatory effects of incoming stimuli
~\cite{Klimesch2007, Wang1996}.

\begin{table*}[!b]
\begin{center}
\begin{threeparttable}
\caption{
\label{table:1}
Hyperparameters of the model. The first column shows the default parameters
  chosen to solve Boolean logic problems. The second column shows the search
  range used in the hyperparameter search. Asterisks ($^*$) mark ranges that
  were probed according to a logarithmic scale; all others were probed
  linearly. The last column shows the value chosen from these ranges to solve
  MNIST.}
\begin{tabular}{p{4.8cm} p{4cm} p{3cm} p{3cm}}
\toprule
\textbf{Parameter} & \textbf{Default value (Boolean tasks)} & \textbf{Search range} & \textbf{Chosen value (MNIST)}\\
\midrule
\texttt{batch\_size}                & $1$           & $[1, 1000]$\tnote{*}                  & $5$ \\
\texttt{clip\_derivative}           & $100.0$       & $[1, 1000]$                           & $539.7$ \\
\texttt{decay\_constant} ($\tau$)   & $1.0$         & $[0.1, 2]$                            & $0.181769$ \\
\texttt{fire\_threshold} ($\theta$) & $1.0$         & $[0.1, 1.5]$                          & $1.16732$ \\
\texttt{learning\_rate}             & $0.001$       & $[10^{-5}, 1.0]$\tnote{*}             & $10^{-4} \times 2.01864$ \\
\texttt{learning\_rate\_pulses}     & $0.001$       & $[10^{-5}, 1.0]$\tnote{*}             & $10^{-2} \times 5.95375$ \\
\texttt{n\_hidden}                  & $1 \times 2$  & $[0, 4] \times [2, 1000]$\tnote{*}    & $1 \times 340$ \\
\texttt{n\_pulses}                  & $1$           & $[0, 10]$                             & $10$ \\
\texttt{nonpulse\_init\_multiplier} & $0.0$         & $[-10, 10]$                           & $-0.275419$ \\
\texttt{penalty\_no\_spike}         & $1.0$         & $[0, 100]$                            & $48.3748$ \\
\texttt{pulse\_init\_multiplier}    & $0.0$         & $[-10, 10]$                           & $7.83912$ \\
\bottomrule
\end{tabular}
\begin{flushright}
* - logarithmic search space
\end{flushright}
\end{threeparttable}
\end{center}
\end{table*}

Despite careful initialisation, the network might still become quiescent during
training. We prevent this problem by adding a fixed small penalty
(\texttt{penalty\_no\_spike}) to the derivative of all presynaptic weights of a
neuron that has not fired. In practice, after the training phase, some of the
neurons will spike too late to matter in the classification and thus they do
need to spike at all.

Another problem is that the gradients become very large as a spike becomes
closer to, but not sufficient for the postsynaptic neuron to reach the firing
threshold. In this case, in Eq.~\ref{eq:5}~and~\ref{eq:6}, the value of the
Lambert $W$ function will approach its minimum ($-1$) as its argument
approaches $-e^{-1}$, the denominator of the derivatives will approach zero and
the derivatives will approach infinity. To counter this, we clip the
derivatives to a fixed value \texttt{clip\_derivative}. Note that this
behaviour will occur in any activation function that has a maximum (hence, a
biologically-plausible shape), is differentiable, and has a continuous
derivative.

In addition to these hyperparameters, we explored several other heuristics for
the spiking net. These included weight decay, adding random noise during
training to the spike times of either the inputs or all non-output neurons in
the network, averaging over brightness values in a convolutional-like manner
and adding additional input neurons responding to the inverted version of the
image, akin to the on/off bipolar cells in the retina. Additionally, to improve
the computation time of the network, we tried removing presynaptic neurons from
the presynaptic set once their individual contribution to the potential decayed
below a decay threshold. This can be achieved by solving an equation similar to
Eq. \ref{eq:2} for reaching a decay threshold on the decaying part of the
function, using the $-1$ branch of the Lambert $W$ function. None of these
techniques improved our results, so we did not include them here.

\subsection{Experiments}

\subsubsection{Boolean logic problems}

We first tested the problem on noisy Boolean logic problems: AND, OR, XOR. For
each example, we encoded the two inputs as the individual spike times of the
two input neurons. All spikes occurred at times between $0$ and $1$. We drew
True and False values from uniform distributions between $[0.0, 0.45]$ and
$[0.55, 1.0]$, respectively. The result is encoded as the first neuron to spike
in the output layer. We used a single synchronisation pulse connected to all
non-input neurons of the network.

We also solved a concentric circles problem, where the value represents a 2D
coordinate uniformly drawn from either an inner circle with a radius of $0.3$
or an outer circle with an inner radius of $0.4$ and an outer radius of $0.5$.

\subsubsection{Non-convolutional MNIST}

\begin{figure}[h]
    \centering
    \includegraphics[width=0.4\textwidth]{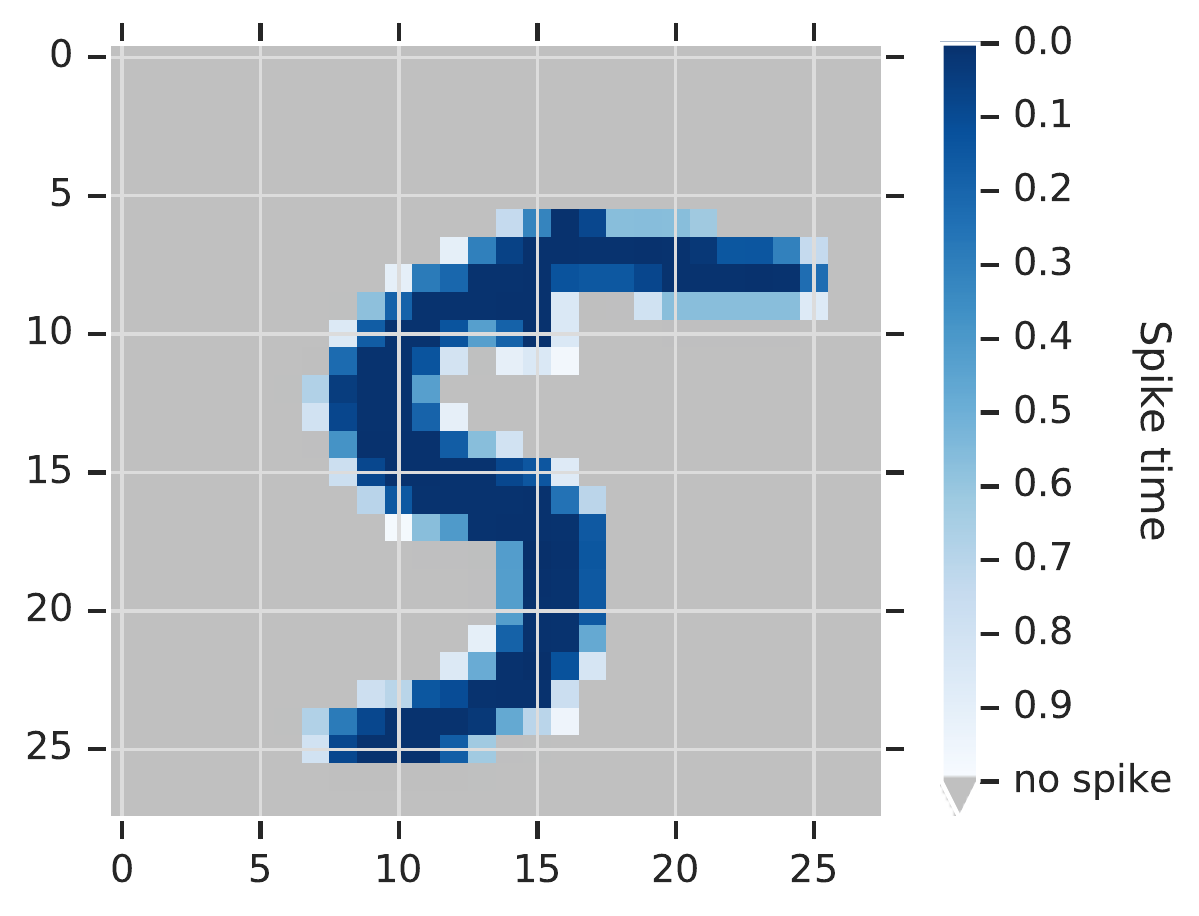}
    \caption{\label{fig:2}Example of an MNIST digit temporal representation.
    Each of the $784$ input neurons spikes at a time proportional with the
    brightness of the corresponding pixel in the flattened row-major brightness
    matrix.}
\end{figure}

We solved the MNIST benchmark by encoding the raw pixel brightness
values in the temporal delay of the $784$ neurons of the input layer
(Figure~\ref{fig:2}). All values were encoded as spikes occurring in the
interval $(0, 1)$. Darker pixels were encoded as earlier spikes compared to
brighter pixels, as they represent more salient information. The scale of the
brightness-to-temporal encoding was linear. Input neurons corresponding to
white pixels did not spike.

The output of the network was encoded as the first neuron to spike among the
$10$ neurons in the output layer. We used individual sets of pulses with fixed
size connected to each individual layer, in order to allow for deeper
architectures where the spike times of different layers might become
considerably different.

To find the optimal parameters for solving MNIST, we performed a hyperparameter
search using evo-lutionary-neural hybrid agents~\cite{Maziarz2018}. This
technique combines deep reinforcement learning with the method of evolutionary
agents and has been shown to outperform both approaches in architecture search
for a variety of classification tasks. It consists of an evolutionary agent
encoding an architecture whose mutations are guided by a neural network trained
with reinforcement learning. We used Google Vizier as infrastructure for the
hyperparameter search~\cite{Golovin2017}.

As detailed in Table \ref{table:1}, the hyperparameter search includes the
decay constant of the neurons, whereas the spike times of the inputs are fixed.
In this work, we make no claims about the biological meaning of the temporal
scales themselves, in particular the spike times of the encoded problem
relatively to the decay constant of the model. Rather, we encode well-known
problems in time in order to show the general capability of spiking models with
temporal coding to solve non-linear, complex problems.

We ran the hyperparameter search with parameter ranges detailed in
Table~\ref{table:1}. Each trial trained a network on a set of parameters
suggested by the hyperparameter search agent within the ranges specified in
Table~\ref{table:1}. The $60000$ MNIST training examples were randomly split
into a training set ($90\%$) and a validation set ($10\%$). The final reported
objective to be maximised by the search agent was the accuracy on the
validation set after 100 epochs. The search was run for $3394$ trials. We then
chose the hyperparameters of the model as the set that produced the best
objective value during the search. These values are given in
Table~\ref{table:1}.

Finally, we used this set of hyperparameters to train $3$ networks for $1000$
epochs. This time we trained on the whole MNIST training data with no
validation set and tested on the MNIST test data.

\subsection{Feature visualisation}

We explored the representations learned by the spiking network using the family
of ``deep dream'' methods as commonly applied in atemporal neural
networks~\cite{Erhan2009, Mordvintsev2015, Olah2017}. Given a network with
fixed weights and pulses, a blank input image can be gradually adjusted in
order to minimise the spike time of a particular neuron. Optionally, at the
same time, the spike times of the other neurons in the same layer can be
maximised. To do this, a one-hot encoded vector is set as the target for a
non-input layer of the network and the derivative with respect to the spike
time (Eq.~\ref{eq:5}) is used to backpropagate errors to the input layer. Then,
the same procedure is repeated with the adjusted image, until the output is
close enough to the target.

We performed feature visualisation on the network that performed best. We
started from a blank image with each pixel initialised to spike at $t = 0.0$,
though we confirmed that starting from a random image is also feasible for this
procedure. We gradually adjusted the input image once per epoch with a learning
rate of $0.1$, until the classification produced the correct target class for
$10$ consecutive epochs. As an additional constraint, we only allowed
non-negative spike times.

\section{Results}

\subsection{Boolean logic problems}

We first trained small networks containing $2$ hidden neurons and $1$
synchronisation pulse to solve 2-di-mensional Boolean logic problems encoded in
spike times. The networks were trained using the default parameters from
Table~\ref{table:1} for a maximum of $100$ epochs on $1000$ training examples.
They were tested on $150$ randomly generated test examples from the same
distribution. We found that we achieved faster training for these problems when
Adam updates were computed and applied only on examples that were classified
wrong. In practice, this method has the advantage of discouraging overfitting
and reducing the number of operations required for training.

\begin{figure*}[t]
    \centering
    \includegraphics[width=0.24\textwidth]{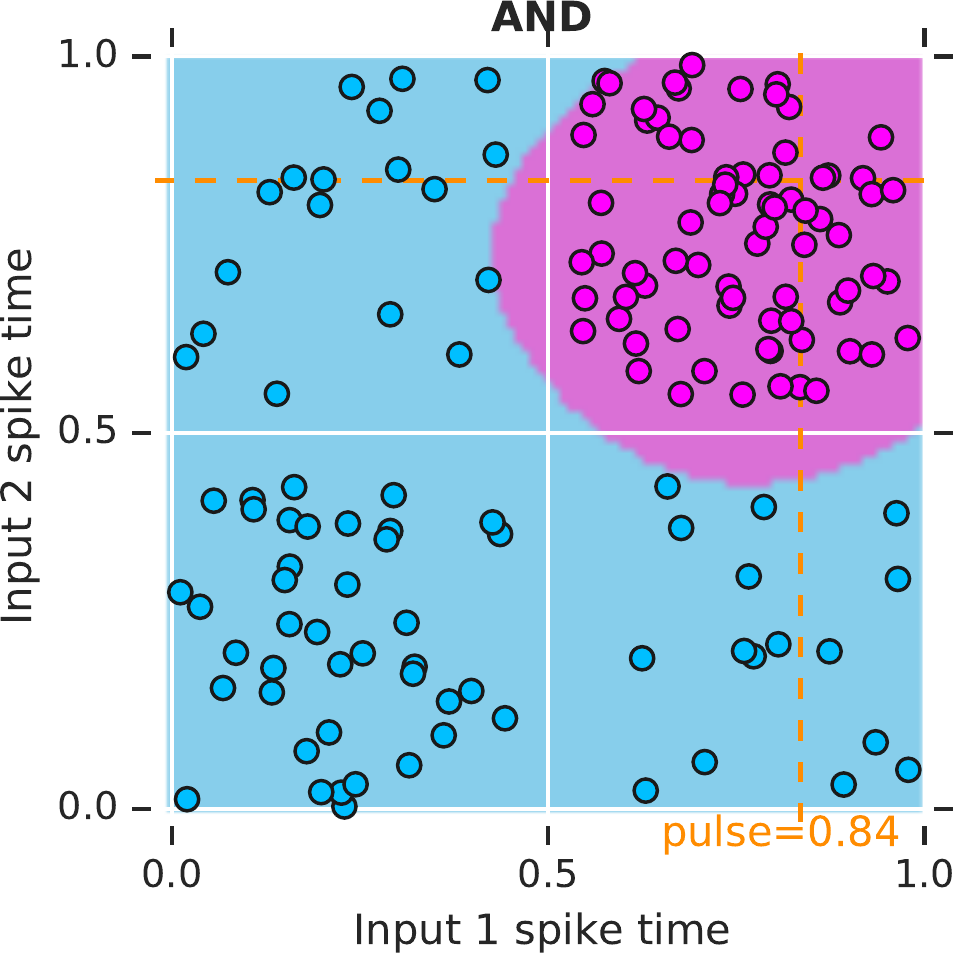}
    \includegraphics[width=0.24\textwidth]{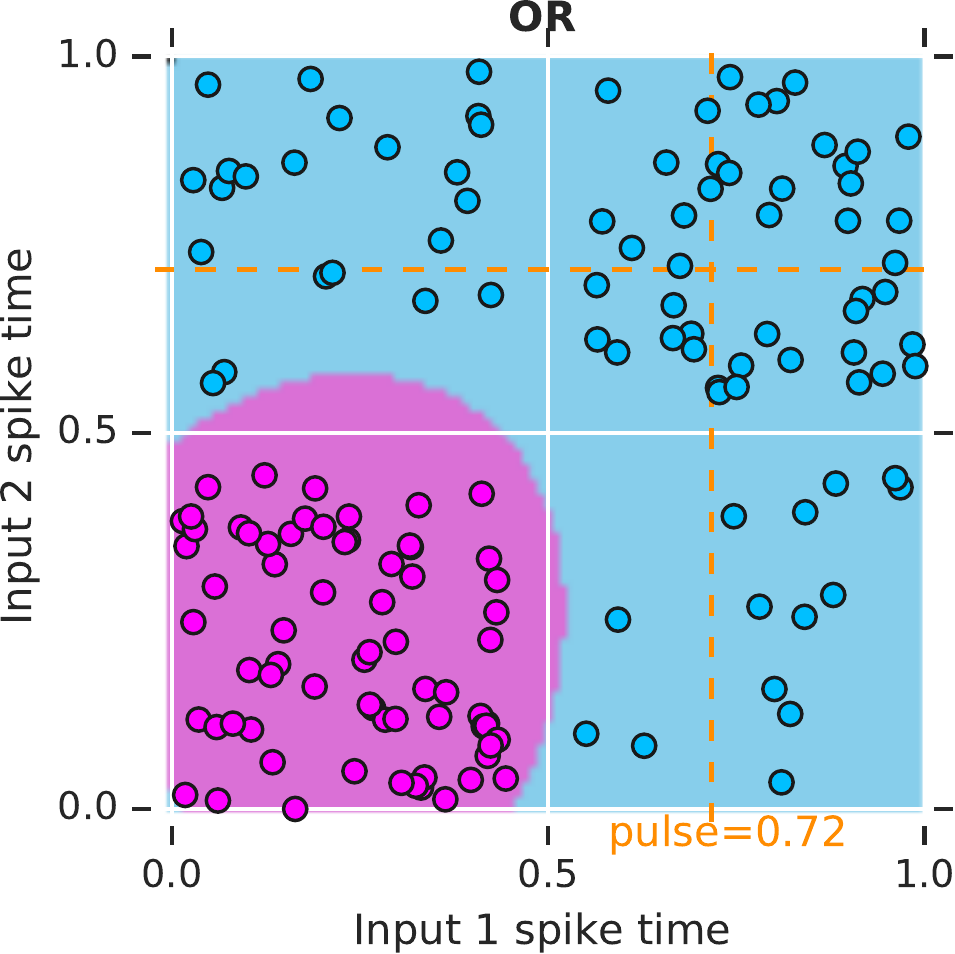}
    \includegraphics[width=0.24\textwidth]{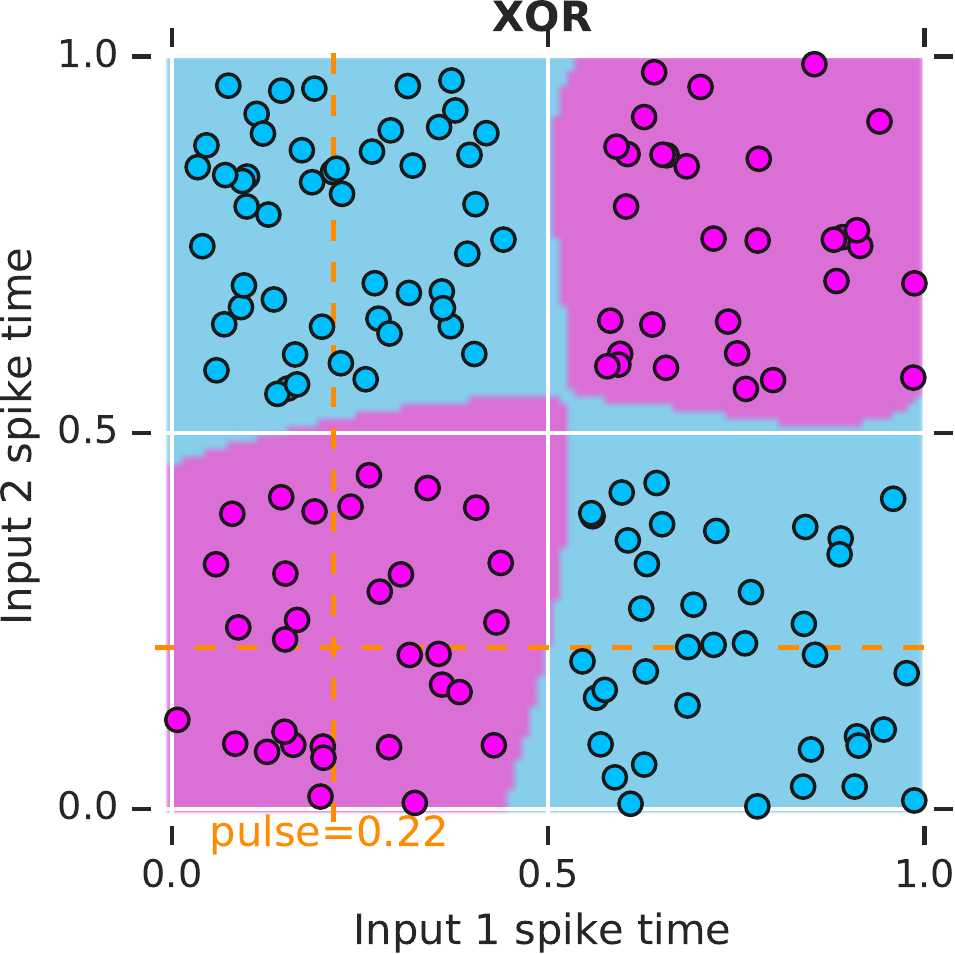}
    \includegraphics[width=0.24\textwidth]{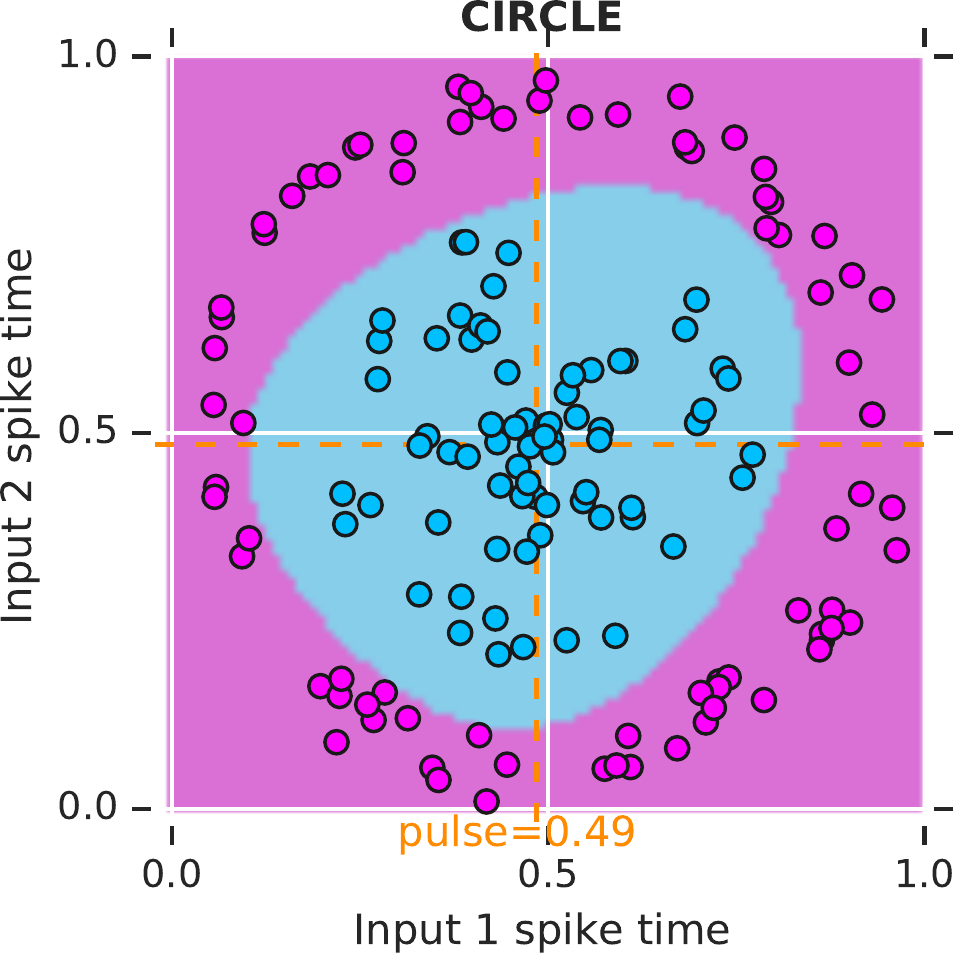}
    \caption{\label{fig:3}Example of temporal class boundaries produced by
    small spiking networks with $2$ hidden neurons and $1$ synchronisation
    pulse connected to all non-input neurons. All networks are trained for a
    maximum of $100$ epochs on $1000$ training examples, while $150$ test
    examples are shown in the figure. The default hyperparameters used for
    training are shown in Table~\ref{table:1}.}
\end{figure*}

We were able to train such small spiking networks with $100\%$ accuracy on all
four problems. Figure~\ref{fig:3} exemplifies class boundaries for such
networks. All trained networks in these examples had only positive weights.

\begin{figure}[!b]
    \centering
    \includegraphics[width=0.5\textwidth]{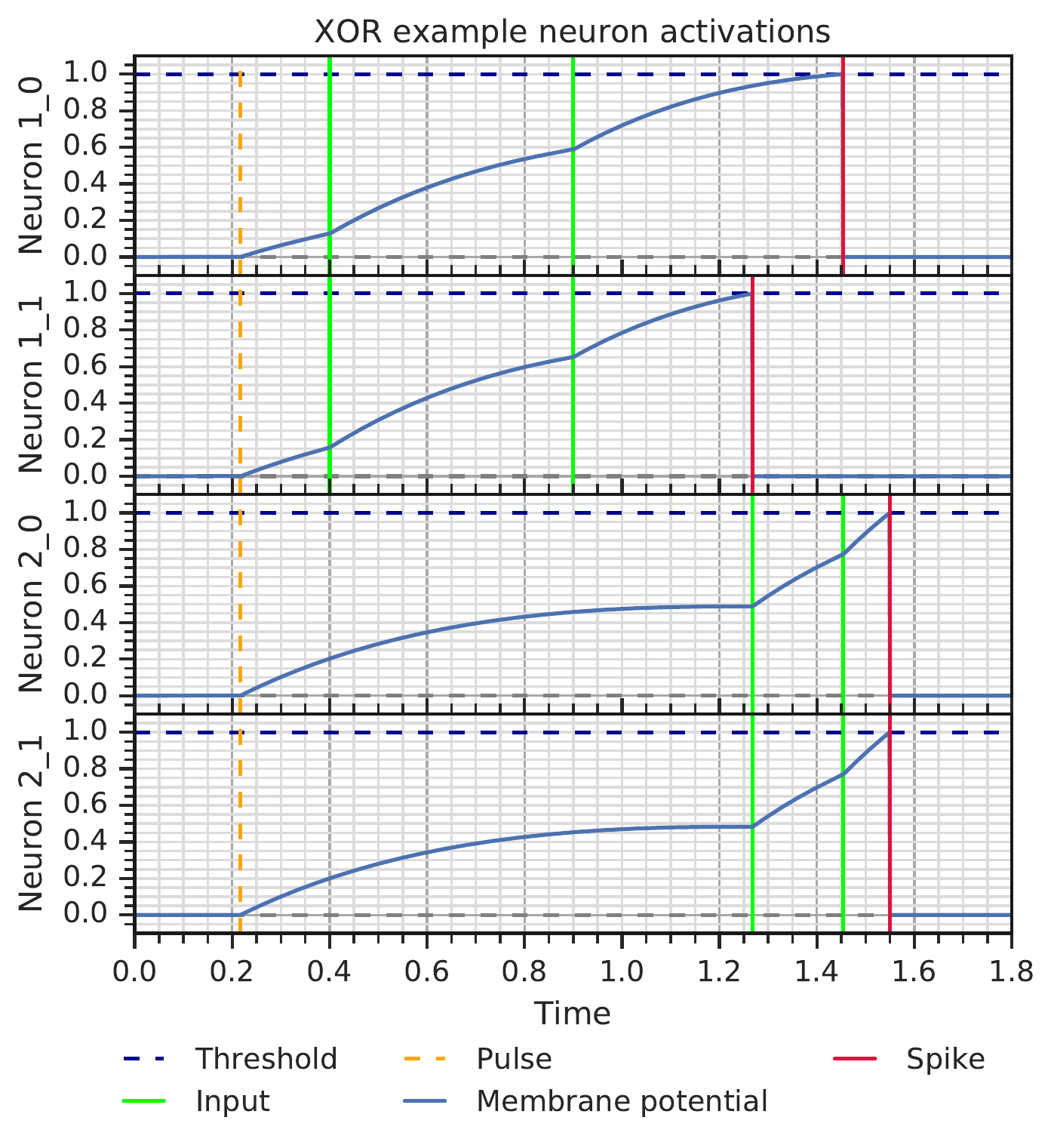}
    \caption{\label{fig:4}Membrane potentials of hidden $(1\_*)$ and output
    $(2\_*)$ neurons during the classification of one XOR example, with
    incoming inputs at $t = 0.4$ and $t = 0.6$.}
\end{figure}

Figure~\ref{fig:4} shows the dynamics of the membrane potential of every
individual neuron during the classification of a typical XOR example. In this
particular case, all neurons spike. With larger spacing between inputs, one of
the hidden neurons no longer needs to spike. Given the limited range of inputs
with respect to the decay constant, the classification occurs before any
significant decay has occurred. However, the network can also be successfully
trained with scaled inputs or decay constant so that the time to reach the
alpha function peak is exceeded during the classification, in which case the
information will also propagate during the decay phase (see
Figure~\ref{fig:1A}).

The purpose of solving a set of simple problems was to demonstrate that spiking
networks of a small size can be trained without hyperparameter tuning, and
hence we did not focus on performance metrics for these problems.

\subsection{Non-convolutional MNIST}

We found that the best out of the three networks obtained as described in the
Methods section reached a maximum accuracy of $99.96\%$ and $97.96\%$ on the
MNIST train and test sets, respectively (Table~\ref{table:2}). The
hyperparameters used to achieve this result are given in Table~\ref{table:1}.
The hyperparameter values are given in the table up to $6$-decimal precision,
however in practice an amount of imprecision can be tolerated. In general, the
best candidate sets of hyperparameters had a low batch size (up to $5$), a
small decay rate (between $0.1$ and $0.3$), and a non-pulse learning rate
usually under $10^{-3}$, whereas the other hyperparameters had comparatively
more varied values.

\begin{table}[h]
\begin{center}
\begin{threeparttable}
\caption{
\label{table:2}
Accuracy of the spiking network trained with the best set of hyperparameters
  during the slow and fast regimes that occurred spontaneously during
  training.}
\begin{tabular}{l l l}
\toprule
 & \textbf{Slow regime} & \textbf{Fast regime}\\
\midrule
\textbf{Training accuracy (\%)}  &$99.9633$      &$99.885$ \\
\textbf{Training loss (mean)}   &$0.002884$     &$0.00444$ \\
\textbf{Test accuracy (\%)}      &$97.96$        &$97.4$ \\
\textbf{Test loss (mean)}       &$0.173248$     &$0.19768$ \\
\bottomrule
\end{tabular}
\end{threeparttable}
\end{center}
\end{table}

This result improves on recent spiking models published recently (but not
trained with conversion from atemporal networks), as detailed in the
Discussion. For comparison, we also trained a fully-connected conventional
multilayer perceptron with ReLU neurons with the same architecture as the
spiking network ($784$--$340$--$10$ neurons) on MNIST. We used the same
training setting as for the spiking network: Glorot initialisation, Adam
optimizer (learning rate $0.001$), the same train/test split, the same number
of epochs. We tested batch sizes $5$, $32$ and $128$. For each batch size, we
ran the training three times.  During the nine runs, we obtained a maximal
accuracy of $97.9\%$ on the test set. Although it can be claimed that various
techniques can improve the performance of conventional neural networks (as it
also may well be the case with the spiking model presented here), we aim to
show that, generally-speaking, the performances of the two types of networks
are comparable.

\begin{figure}[t]
    \centering
    \includegraphics[width=0.5\textwidth]{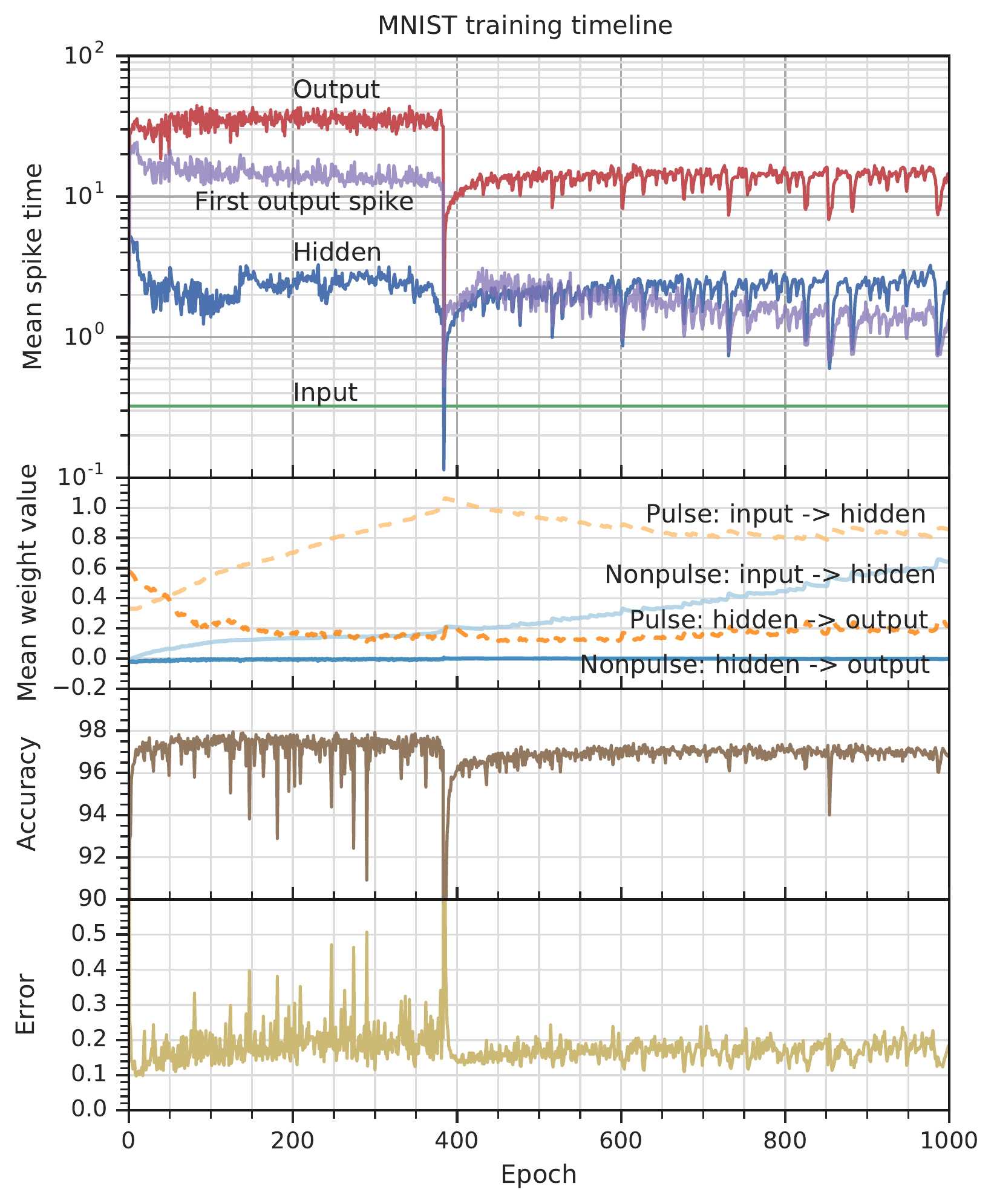}
    \caption{\label{fig:5}Learning dynamics during the training of a spiking
    network with the best set of hyperparameters. The plots show a change in
    regime at epoch $384$ from slow to fast classification. All statistics are
    shown for the test set.}
\end{figure}

During the training process, we observed the same spiking network spontaneously
learning two different operating modes for the classification of inputs: a slow
regime and a fast regime (Figure~\ref{fig:5}). In the slow regime, the network
spikes in a typical feedforward manner, with the neurons in the output layer
usually waiting for all hidden neurons and pulses to spike before spiking
themselves. The best accuracy is reported during the slow regime. On the other
hand, in the fast regime, the network makes very fast decisions, with the first
spike in the output layer occurring before the mean spike in the hidden layer.
The transitions between the two regimes is sudden.

\begin{figure*}[!h]
    \centering
    \begin{subfigure}[h]{0.49\textwidth}
        \includegraphics[width=\textwidth]{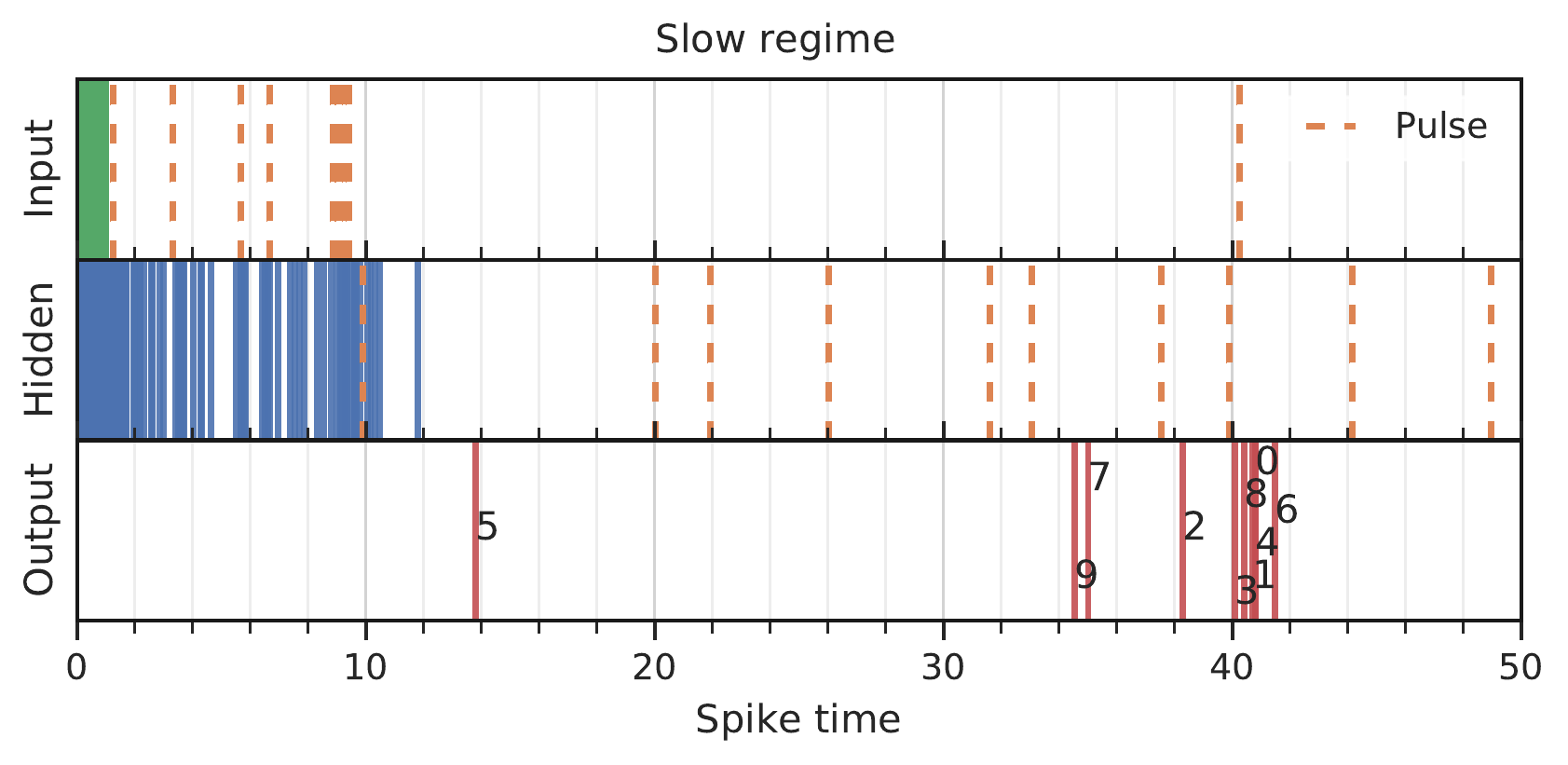}
    \end{subfigure}
    \begin{subfigure}[h]{0.49\textwidth}
        \includegraphics[width=\textwidth]{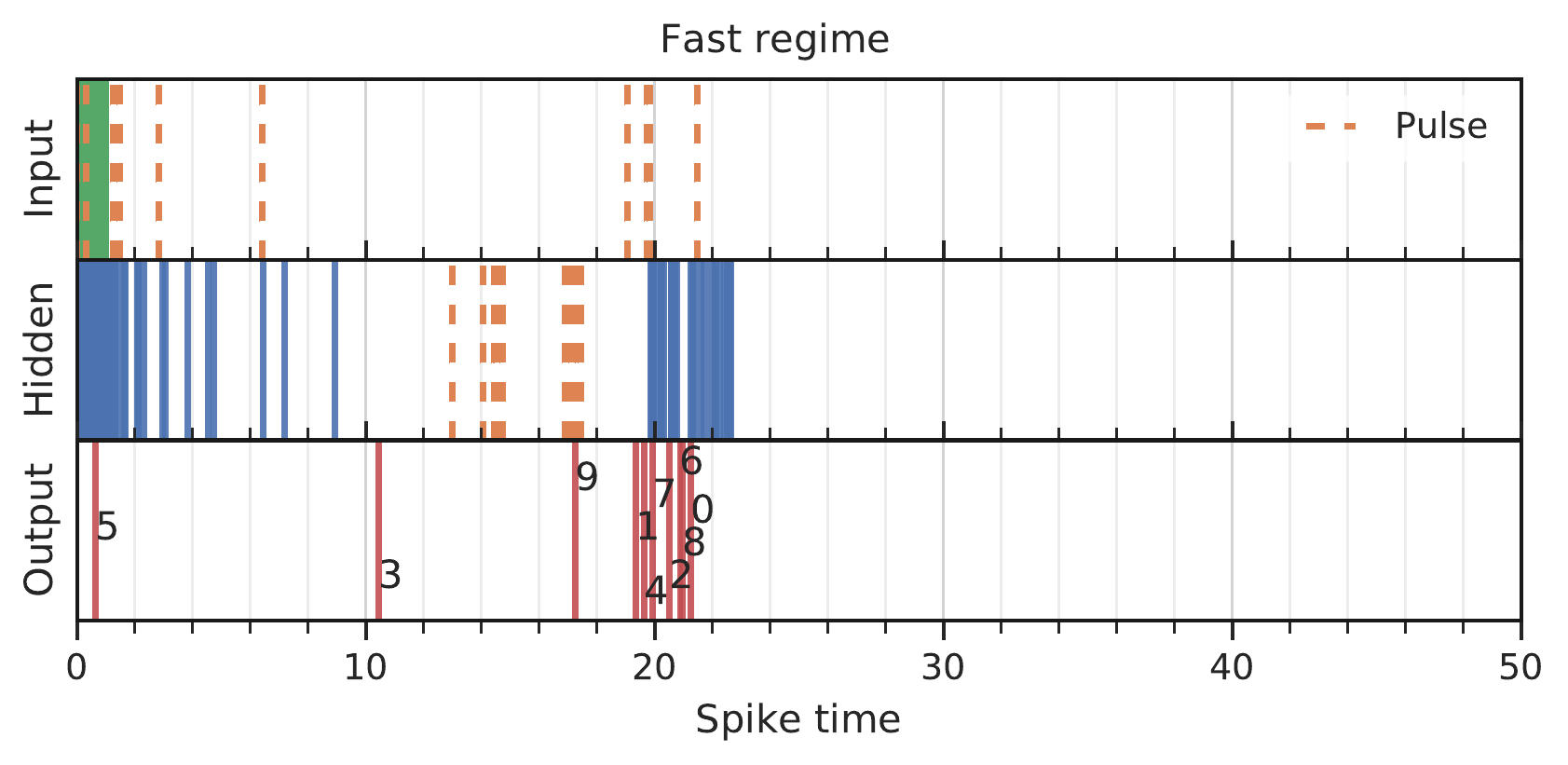}
    \end{subfigure}
    \caption{\label{fig:6}Raster plot of spikes in each layer during the
    classification of the digit shown in Figure~\ref{fig:2} by the
    best-performing networks with slow and fast regimes. The pulses shown in
    each layer are connected to the neurons in the next layer. The number
    labels indicate the digit encoded by an output neuron.}
\end{figure*}

\begin{figure*}[!h]
    \centering
    \includegraphics[width=0.7\textwidth]{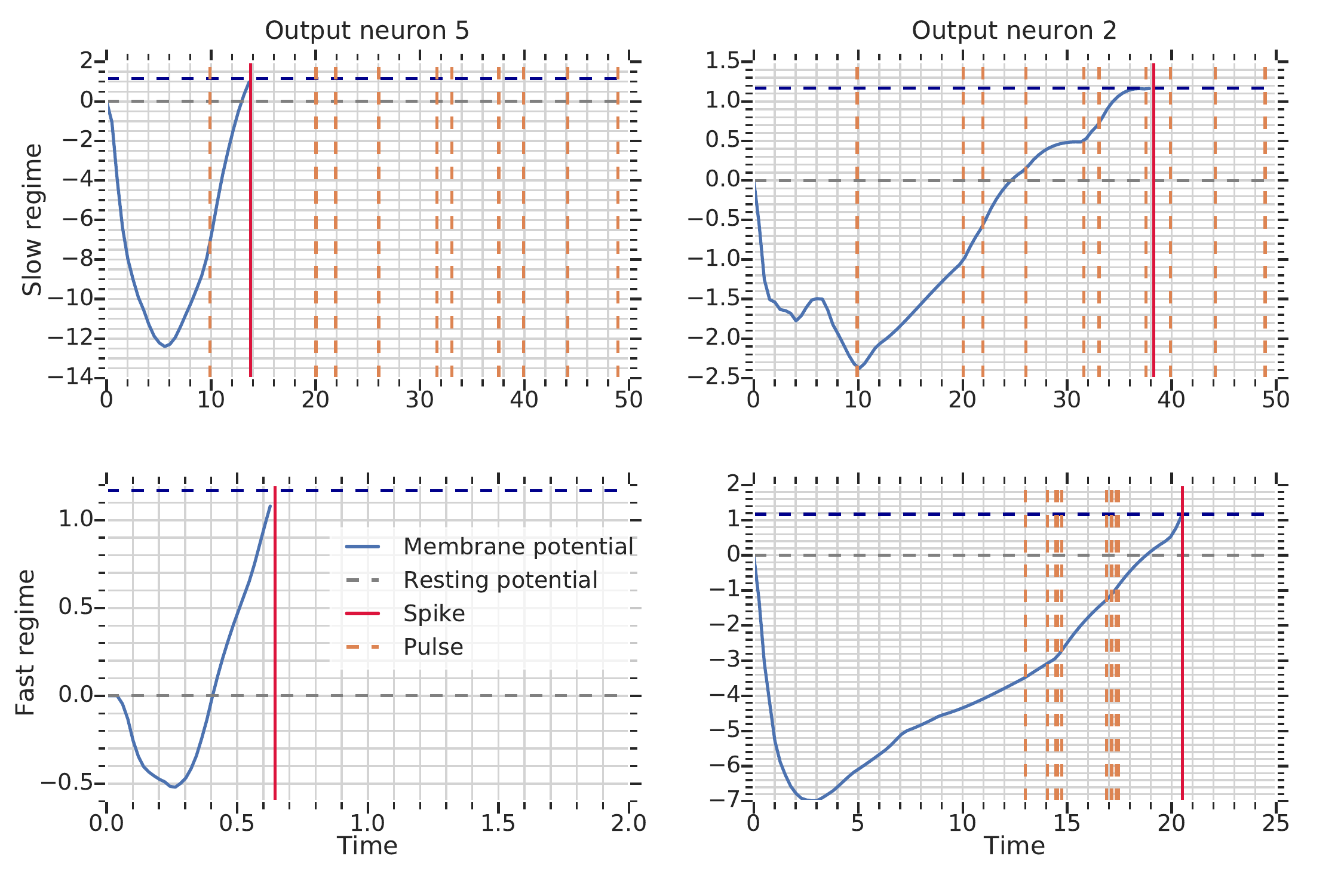}
    \caption{\label{fig:7}Membrane potentials of two output target and
    non-target neurons during the classification of the digit shown in
    Figure~\ref{fig:2} by the best-performing networks with slow and fast
    regimes. The axis scales are different for each plot, reflecting the
    considerable difference in speed of classification between the two regimes.
    Inhibition levels (reflected in negative amplitudes) are also considerably
    different in the two regimes. The membrane potential always becomes
    hyperpolarised (achieves negative potential) in output neurons before they
    spike.}
\end{figure*}

\begin{figure}[!h]
    \centering
    \includegraphics[width=0.5\textwidth]{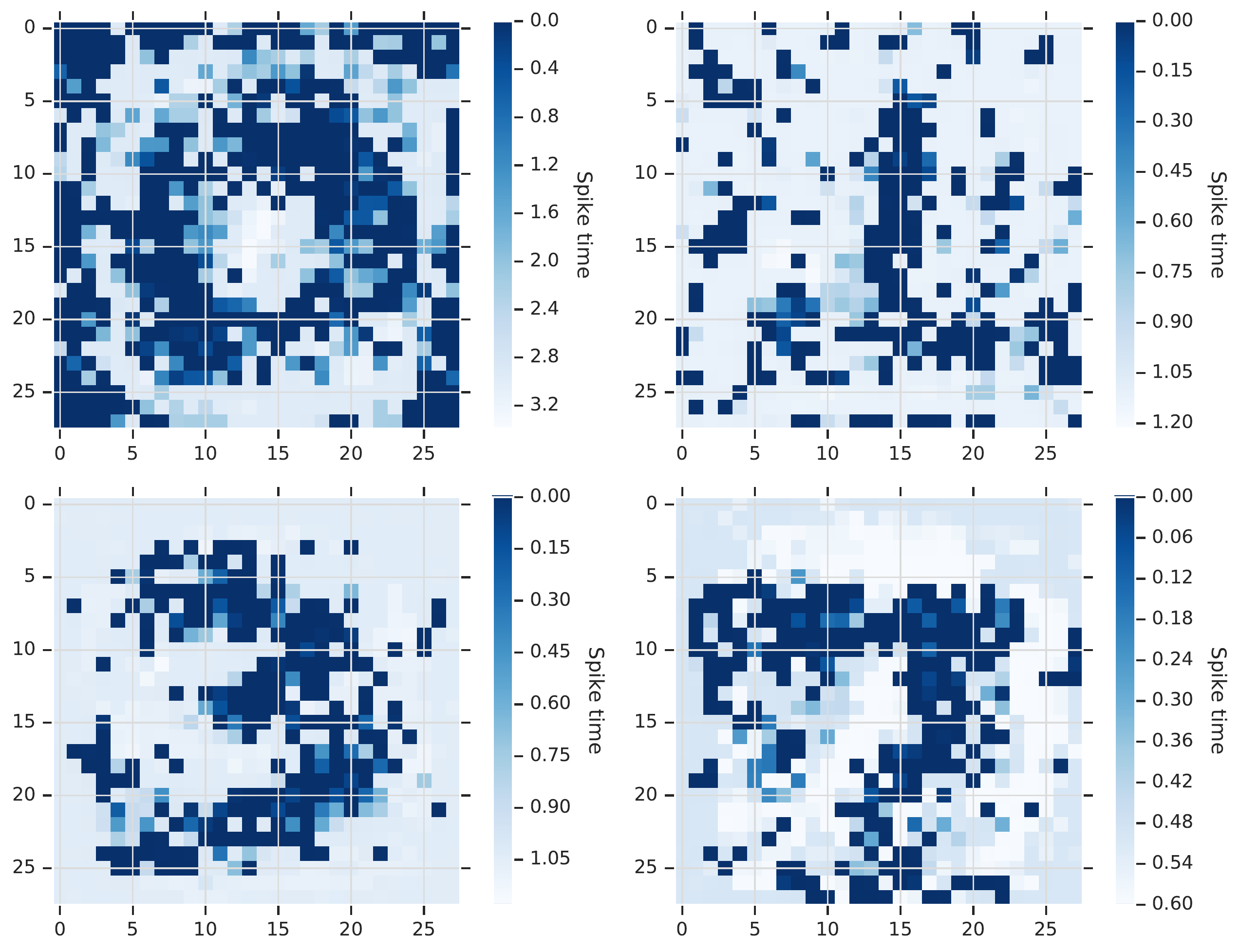}
    \caption{\label{fig:8}Example of reconstructed input maps that produce a
    particular output class in the best-performing slow regime network. The
    target classes are, in row-major order, $0$, $1$, $3$ and $7$. The start
    image is filled with zeros and adjusted using standard gradient descent
    until the target class is produced $10$ epochs in a row. Only non-negative
    spike times are allowed.}
\end{figure}

Investigating this transition, we found that the input layer pulses had an
oscillatory behaviour during training (likely due to a high learning rate
pre-set using hyperparameter search) and the transition occurred when these
pulses synchronised and simultaneously reached a minimal spike time (0.0).
Although they then recovered to larger non-synchronous spike times, this drove
the hidden layer pulses to spike considerably earlier (Figure~\ref{fig:6}). We
observed the same transition from a slow to a fast regime occurring in one of
the other two trained networks, whereas the third network went directly into a
fast regime. With this set of hyperparameters, in particular the relatively
high learning rate for pulses leading to oscillatory changes across training
epochs, it is likely that the network has a preference for settling into the
fast regime.

In the best-performing slow and fast networks, we investigated the individual
membrane potentials of the output neurons during the classification of
individual examples (Figure~\ref{fig:7}). We found that all output neurons were
first inhibited before producing a spike, even in the case of the fast network.
Moreover, in both networks, the winning neuron spiked before most or all of the
hidden layer pulses spiked. This suggests that one role of the pulses was to
produce late spikes where neurons would not have normally spiked, and thus to
allow the flow of gradients from non-target neurons during training.

The optimal value for the decay constant ($0.181769$) chosen by the
hyperparameter search means that the alpha function has its maximum at
$t=5.56$. The input spikes were prefixed to occur between $0$ and $1$. On the
other hand, the pulses learned to produce spikes at much later times (Figure
6). This means that the decaying part of the function was exploited indeed, as
the output neurons generally spiked only after a subset of the pulses.

Finally, we investigated whether it was possible to reconstruct target digits
using the ``deep dream'' method. It was indeed possible to produce recognizable
digits in this manner (Figure~\ref{fig:8}).

\section{Discussion}

In this paper, we proposed a spiking neural network with biologically-plausible
alpha synaptic function that encodes information in the relative timing of
individual spikes. The activation function is differentiable with respect to
time when the presynaptic set is fixed, thus making the network trainable using
gradient descent. A set of synchronisation pulses is used to provide bias and
extra degrees of freedom during training. We showed that the network is able to
successfully solve benchmark problems encoded in time, such as noisy Boolean
logic problems and MNIST. Our open-source code and network is openly available
at~\cite{ihmehimmeli2019}.

Our model improves upon existing spiking network models, in particular those
from the SpikeProp family, in multiple ways. First, we use a
biologically-plausible alpha synaptic function. The alpha function includes a
decay component that allows earlier inputs to be forgotten when a neuron has
not spiked, which can help in correcting potentially false alarms. We also use
a continuous (as opposed to a discretised) scheme where the spike time of an
input neuron is proportional to the value of a feature, such as the brightness
of a pixel. Finally, we use trainable synchronisation pulses connected to
individual layers of the network, which provide temporal bias and allow extra
flexibility during training,

Our results on the MNIST benchmark compare favourably with other recent models
from literature, including similar models with larger or deeper architectures.
Recent works on non-convolutional spiking network whose benchmark results we
improve include~\cite{Mostafa2017, OConnor2016, Diehl2016, Tavanaei2019bpstdp,
Querlioz2013, Brader2007}. Recently, performant spiking networks have been
obtained by conversion from atemporal neural networks~\cite{Tavanaei2019deep},
however we are not aware of such networks performing computations by encoding
information in the timing of individual spikes.

The main challenge posed by this type of model is the discontinuous nature of
the loss function at points where an output neuron stops spiking and due to the
landscape changes at the points where the order of pre- and postsynaptic spikes
changes. In general, discontinuities also appear in other deep learning models,
such as the derivative of the ReLU activation function in atemporal networks.
We found that the training process is able to overcome these challenges.

Another practical challenge is the computational complexity of the activation
function and the training process. In conventional atemporal networks, the
feedforward pass can be performed as a sequence of matrix multiplications,
which can be efficiently parallelised on the GPU or other specialised hardware.
In spiking networks, the feedforward pass cannot be parallelised in the same
efficient fashion, leading to slower training times. In addition, solving the
threshold crossing equation in the spiking network requires the computation of
exponentials and solving the Lambert $W$ function, which is relatively
expensive. The networks presented in this work are trainable on the order of a
few seconds to a few hours.

Despite challenges, there are two prominent goals motivating progress in the
field of spiking neural networks. On the one hand, spiking neural networks are
interesting as computational models of biological processes. They can
contribute to understanding low-level information processing occurring in the
brain and help us in the search for neural correlates of
cognition~\cite{Ward2003, Shanahan2008, Gamez2009}. On the other hand, they can
be deployed in neuromorphic hardware~\cite{Davies2018, Pfeiffer2018, Neftci2017}
to implement rapid and energy-efficient computations.

From a computational biology perspective, the main contribution of this work is
showing that it is possible to perform complex, nonlinear computations in
networks with biologically-plausible activation functions that encode
information in the timing of single spikes. The timing of neuronal spikes, as
opposed to spike rates, must play an important role in biological brains during
the fast processing of perceptual stimuli. Neurons involved in the processing
of sensory stimuli, such as ganglion cells in the retina, fire with very high
precision and convey more information through their timing then through their
spike count~\cite{Berry1997}. The relative difference between the first spikes
of different ganglion cells can encode the spatial structure of an
image~\cite{Gollisch2008}. In the primary sensory neurons in human fingertips,
the relative timing of the first spikes encodes information about tactile
pressure and texture~\cite{Johansson2004}. Further constraints can be inferred
from the response times of cortical neurons. In macaque brains, neurons in the
temporal area can respond to a visual stimulus within 100ms; given that the
anatomy of the visual cortex indicates that at least 10 synaptic stages must be
passed to reach the temporal cortex, and given the conduction speed of
intracortical fibers and the distance between the brain areas involved, it must
be that the response is generated by single, or at most double, spikes in
individual neurons~\cite{Thorpe1989}. It must therefore be that the temporal
encoding of stimuli into single spikes fired by individual neurons are
essential in rapid stimulus processing.

More generally, relative spike times are reflected in phase synchronisation
between neuronal populations, which can be observed both in individual-neuron
studies~\cite{Moiseff1981} and in macroscale recording of brain
activity~\cite{Palva2016}. Such synchronisation has been shown to be reliable
across different neural stages~\cite{Moiseff1981, Diesmann1999, Reinagel2000,
Huxter2003}, particularly in dynamic contexts~\cite{Lestienne2001}. The
relative encoding of information in the precise timing of spikes produced by
single neurons is thought to represent a solution to the binding problem ---
the ability of the brain to coordinate over features highly distributed neural
representations, such as associating a colour and a shape with a particular
object in the environment~\cite{Engel1992, Gray1999}.

We do not claim that temporal coding is solely responsible for all of the
complex cognitive processes performed by biological brains, but rather aim to
draw attention that this type of encoding can be an efficient and
computationally powerful alternative to rate coding. A reasonable hypothesis is
that the two encoding methods are complementary in the brain. A temporal code
can be transformed into a rate codes~\cite{Ahissar1998} and vice
versa~\cite{Mehta2002}.

Another point that deserves discussion is backpropagation in biological
networks. Our investigation was concerned with the representational capacity,
as opposed to the learning mechanism of the network. We therefore used
backpropagation to teach the spiking network. The idea of backpropagation in
neural networks has for a long time been considered
implausible~\cite{Crick1989}, due to requirements such as the existence of
symmetrical connections between neuronal layers. However, recent works have
begun to challenge this viewpoint and have made promising attempts to
understand how backpropagation-like mechanisms might work in a biological
network~\cite{Hassabis2017, Liao2015, Bengio2014, Lillicrap2016,
Whittington2017}. Notably, it has been shown that random feedback connections
are able to support learning with backpropagation~\cite{Lillicrap2016}. The
main learning mechanism observed in biological neurons is spike-time dependent
plasticity~\cite{Song2000}, which can theoretically implement weight updates as
required by backpropagation~\cite{Bengio2015, Whittington2017}. This topic is
nevertheless still an open question. Neuroscience is still uncovering the role
of neural elements previously thought to have passive roles in the nervous
system, which in fact appear to actively participate in synaptic regulation;
they might also have significant roles for learning in the
brain~\cite{Newman2003, Eroglu2010}.

To conclude, here we showed that a spiking neural network model with
biologically-plausible characteristics is able to solve standard machine
learning benchmark problems using temporal coding. These results invite further
exploration into the possibility of using such networks to generate behaviours
resembling those of biological organisms, as well as investigations into the
computational capabilities of spiking neural networks with more complex and
recurrent architectures.

On a more speculative note, we expect that further advances in spiking networks
with temporal coding will open exciting new possibilities for artificial
intelligence, in the same way that it has for biological brains. We envision
that neural spike-based state machines will offer a natural solution for the
efficient modelling and processing of real-world analogue signals. It will be
possible to integrate artificial spiking networks with biological neural
networks and create interfaces between the two. These advances will come with
significant computing energy savings. By observing and testing models such as
the one presented in this study, we aim to increase the familiarity of the
research community with the temporal coding paradigm and to create building
blocks towards recurrent and state-based architectures for neural computing.

\bibliography{references}

\begin{thebibliography}{10}

\bibitem{Silver2016}
D.~Silver, A.~Huang, C.~J. Maddison, A.~Guez, L.~Sifre, G.~van~den Driessche,
  J.~Schrittwieser, I.~Antonoglou, V.~Panneershelvam, M.~Lanctot, S.~Dieleman,
  D.~Grewe, J.~Nham, N.~Kalchbrenner, I.~Sutskever, T.~Lillicrap, M.~Leach,
  K.~Kavukcuoglu, T.~Graepel, and D.~Hassabis, ``{Mastering the game of Go with
  deep neural networks and tree search},'' {\em Nature}, vol.~529,
  pp.~484--489, jan 2016.

\bibitem{Krizhevsky2012}
A.~Krizhevsky, I.~Sutskever, and G.~E. Hinton, ``{ImageNet classification with
  deep convolutional neural networks},'' {\em Communications of the ACM},
  vol.~60, pp.~84--90, may 2017.

\bibitem{Szegedy2015}
C.~Szegedy, {Wei Liu}, {Yangqing Jia}, P.~Sermanet, S.~Reed, D.~Anguelov,
  D.~Erhan, V.~Vanhoucke, and A.~Rabinovich, ``{Going deeper with
  convolutions},'' in {\em 2015 IEEE Conference on Computer Vision and Pattern
  Recognition (CVPR)}, pp.~1--9, IEEE, jun 2015.

\bibitem{Gollisch2008}
T.~Gollisch and M.~Meister, ``{Rapid neural coding in the retina with relative
  spike latencies},'' {\em Science}, vol.~319, no.~5866, pp.~1108--1111, 2008.

\bibitem{Moiseff1981}
A.~Moiseff and M.~Konishi, ``{Neuronal and behavioral sensitivity to binaural
  time differences in the owl},'' {\em The Journal of Neuroscience}, vol.~1,
  pp.~40--48, jan 1981.

\bibitem{Johansson2004}
R.~S. Johansson and I.~Birznieks, ``{First spikes in ensembles of human tactile
  afferents code complex spatial fingertip events},'' {\em Nature
  Neuroscience}, vol.~7, pp.~170--177, feb 2004.

\bibitem{Reinagel2000}
P.~Reinagel and R.~C. Reid, ``{Temporal coding of visual information in the
  thalamus.},'' {\em The Journal of Neuroscience}, vol.~20, pp.~5392--400, jul
  2000.

\bibitem{Huxter2003}
J.~Huxter, N.~Burgess, and J.~O'Keefe, ``{Independent rate and temporal coding
  in hippocampal pyramidal cells.},'' {\em Nature}, vol.~425, pp.~828--32, oct
  2003.

\bibitem{Thorpe1989}
S.~J. Thorpe and M.~Imbert, ``{Biological constraints on connectionist
  modelling},'' {\em Connectionism in perspective}, no.~July 2016, pp.~1--36,
  1989.

\bibitem{Hochreiter1997}
S.~Hochreiter and J.~Schmidhuber, ``{Long Short-Term Memory},'' {\em Neural
  Computation}, vol.~9, pp.~1735--1780, nov 1997.

\bibitem{Speelpenning1980}
B.~Speelpenning, {\em {Compiling fast partial derivatives of functions given by
  algorithms}}.
\newblock PhD thesis, Illinois University, Urbana (USA), 1980.

\bibitem{Rummelhart1987}
D.~E. Rumelhart, G.~E. Hinton, and R.~J. Williams, ``{Learning Internal
  Representations by Error Propagation},'' tech. rep., US Dept of the Navy,
  Cambridge, MA, sep 1985.

\bibitem{Wu2018}
Y.~Wu, L.~Deng, G.~Li, J.~Zhu, and L.~Shi, ``{Spatio-Temporal Backpropagation
  for Training High-Performance Spiking Neural Networks},'' {\em Frontiers in
  Neuroscience}, vol.~12, pp.~1--12, may 2018.

\bibitem{Bellec2018}
G.~Bellec, D.~Salaj, A.~Subramoney, R.~Legenstein, and W.~Maass, ``{Long
  short-term memory and learning-to-learn in networks of spiking neurons},''
  mar 2018.

\bibitem{Lee2019}
C.~Lee, S.~S. Sarwar, and K.~Roy, ``{Enabling Spike-based Backpropagation in
  State-of-the-art Deep Neural Network Architectures},'' mar 2019.

\bibitem{Sussillo2009}
D.~Sussillo and L.~F. Abbott, ``{Generating coherent patterns of activity from
  chaotic neural networks.},'' {\em Neuron}, vol.~63, pp.~544--57, aug 2009.

\bibitem{Xu2013}
Y.~Xu, X.~Zeng, L.~Han, and J.~Yang, ``{A supervised multi-spike learning
  algorithm based on gradient descent for spiking neural networks},'' {\em
  Neural Networks}, vol.~43, pp.~99--113, jul 2013.

\bibitem{Florian2012}
R.~V. Florian, ``{The chronotron: A neuron that learns to fire temporally
  precise spike patterns},'' {\em PLoS ONE}, vol.~7, no.~8, p.~e40233, 2012.

\bibitem{Gutig2006}
R.~G{\"{u}}tig and H.~Sompolinsky, ``{The tempotron: a neuron that learns spike
  timing–based decisions},'' {\em Nature Neuroscience}, vol.~9, pp.~420--428,
  mar 2006.

\bibitem{Huh2017}
D.~Huh and T.~J. Sejnowski, ``{Gradient Descent for Spiking Neural Networks},''
  {\em Advances in Neural Information Processing Systems 31}, pp.~1433--1443,
  jun 2017.

\bibitem{Zenke2018}
F.~Zenke and S.~Ganguli, ``{SuperSpike: Supervised Learning in Multilayer
  Spiking Neural Networks},'' {\em Neural Computation}, vol.~30,
  pp.~1514--1541, jun 2018.

\bibitem{Neftci2017}
E.~Neftci, C.~Augustine, S.~Paul, and G.~Detorakis, ``{Event-driven random
  backpropagation: Enabling neuromorphic deep learning machines},'' in {\em
  2017 IEEE International Symposium on Circuits and Systems (ISCAS)}, vol.~11,
  pp.~1--4, IEEE, may 2017.

\bibitem{Sporea2013}
I.~Sporea and A.~Gr{\"{u}}ning, ``{Supervised Learning in Multilayer Spiking
  Neural Networks},'' {\em Neural Computation}, vol.~25, pp.~473--509, feb
  2013.

\bibitem{Esser2015}
S.~K. Esser, R.~Appuswamy, P.~Merolla, J.~V. Arthur, and D.~S. Modha,
  ``{Backpropagation for Energy-Efficient Neuromorphic Computing},'' {\em
  Neural Information Processing Systems (NIPS)}, pp.~1117--1125, 2015.

\bibitem{Sengupta2019}
A.~Sengupta, Y.~Ye, R.~Wang, C.~Liu, and K.~Roy, ``{Going Deeper in Spiking
  Neural Networks: VGG and Residual Architectures},'' {\em Frontiers in
  Neuroscience}, vol.~13, pp.~1--16, mar 2019.

\bibitem{Hunsberger2016}
E.~Hunsberger and C.~Eliasmith, ``{Training Spiking Deep Networks for
  Neuromorphic Hardware},'' pp.~1--10, 2016.

\bibitem{Rueckauer2018}
B.~Rueckauer and S.-C. Liu, ``{Conversion of analog to spiking neural networks
  using sparse temporal coding},'' in {\em IEEE International Symposium on
  Circuits and Systems (ISCAS)}, vol.~2018-May, pp.~1--5, IEEE, may 2018.

\bibitem{Diehl2016}
P.~U. Diehl, G.~Zarrella, A.~Cassidy, B.~U. Pedroni, and E.~Neftci,
  ``{Conversion of artificial recurrent neural networks to spiking neural
  networks for low-power neuromorphic hardware},'' in {\em IEEE International
  Conference on Rebooting Computing (ICRC)}, pp.~1--8, IEEE, oct 2016.

\bibitem{Hu2018}
Y.~Hu, H.~Tang, Y.~Wang, and G.~Pan, ``{Spiking Deep Residual Network},'' apr
  2018.

\bibitem{Pavlidis2005}
N.~Pavlidis, O.~Tasoulis, V.~Plagianakos, G.~Nikiforidis, and M.~Vrahatis,
  ``{Spiking neural network training using evolutionary algorithms},'' in {\em
  Proceedings. 2005 IEEE International Joint Conference on Neural Networks,
  2005.}, vol.~4, pp.~2190--2194, IEEE, 2005.

\bibitem{Florian2007}
R.~V. Florian, ``{Reinforcement learning through modulation of
  spike-timing-dependent synaptic plasticity},'' {\em Neural Computation},
  vol.~19, no.~6, pp.~1468--1502, 2007.

\bibitem{Bohte2000}
S.~M. Bohte, H.~{La Poutr{\'{e}}}, and J.~N. Kok, ``{Error-Backpropagation in
  Temporally Encoded Networks of Spiking Neurons},'' {\em Neurocomputing},
  vol.~48, pp.~17--37, 2000.

\bibitem{Hong2017}
C.~Hong, ``{Training Spiking Neural Networks for Cognitive Tasks: A Versatile
  Framework Compatible to Various Temporal Codes},'' 2017.

\bibitem{Schrauwen2004}
B.~Schrauwen and J.~{Van Campenhout}, ``{Extending SpikeProp},'' in {\em IEEE
  International Joint Conference on Neural Networks (IEEE Cat. No.04CH37541)},
  pp.~471--475, IEEE, 2004.

\bibitem{McKennoch2006}
S.~McKennoch, {Dingding Liu}, and L.~Bushnell, ``{Fast Modifications of the
  SpikeProp Algorithm},'' in {\em The 2006 IEEE International Joint Conference
  on Neural Network Proceedings}, pp.~3970--3977, IEEE, 2006.

\bibitem{Booij2005}
O.~Booij and H.~tat Nguyen, ``{A gradient descent rule for spiking neurons
  emitting multiple spikes},'' {\em Information Processing Letters}, vol.~95,
  pp.~552--558, sep 2005.

\bibitem{Mostafa2017}
H.~Mostafa, ``{Supervised Learning Based on Temporal Coding in Spiking Neural
  Networks},'' {\em IEEE Transactions on Neural Networks and Learning Systems},
  pp.~1--9, 2017.

\bibitem{Kheradpisheh2018}
S.~R. Kheradpisheh, M.~Ganjtabesh, S.~J. Thorpe, and T.~Masquelier,
  ``{STDP-based spiking deep convolutional neural networks for object
  recognition},'' {\em Neural Networks}, vol.~99, pp.~56--67, mar 2018.

\bibitem{Mozafari2018}
M.~Mozafari, S.~R. Kheradpisheh, T.~Masquelier, A.~Nowzari-Dalini, and
  M.~Ganjtabesh, ``{First-Spike-Based Visual Categorization Using
  Reward-Modulated STDP},'' {\em IEEE Transactions on Neural Networks and
  Learning Systems}, vol.~29, pp.~6178--6190, dec 2018.

\bibitem{Hodgkin1952}
A.~L. Hodgkin and A.~F. Huxley, ``{A quantitative description of membrane
  current and its application to conduction and excitation in nerve.},'' {\em
  The Journal of physiology}, vol.~117, pp.~500--44, aug 1952.

\bibitem{Gerstner2001}
W.~Gerstner, ``{Chapter 12. A framework for spiking neuron models: The spike
  response model},'' in {\em Handbook of Biological Physics}, vol.~4,
  pp.~469--516, 2001.

\bibitem{Sterratt2018}
D.~Sterratt, B.~Graham, A.~Gillies, and D.~Willshaw, ``{The synapse},'' in {\em
  Principles of Computational Modelling in Neuroscience}, pp.~172--195,
  Cambridge: Cambridge University Press, 2018.

\bibitem{Rall1967}
W.~Rall, ``{Distinguishing theoretical synaptic potentials computed for
  different soma-dendritic distributions of synaptic input.},'' {\em Journal of
  Neurophysiology}, vol.~30, pp.~1138--1168, sep 1967.

\bibitem{Frank1959}
K.~Frank, ``{Basic Mechanisms of Synaptic Transmission in the Central Nervous
  System},'' {\em IRE Transactions on Medical Electronics}, vol.~ME-6,
  pp.~85--88, jun 1959.

\bibitem{Burke1967}
R.~E. Burke, ``{Composite nature of the monosynaptic excitatory postsynaptic
  potential.},'' {\em Journal of Neurophysiology}, vol.~30, pp.~1114--1137, sep
  1967.

\bibitem{Maziarz2018}
K.~Maziarz, M.~Tan, A.~Khorlin, K.-Y.~S. Chang, S.~Jastrz{\c{e}}bski,
  Q.~de~Laroussilhe, and A.~Gesmundo, ``{Evolutionary-Neural Hybrid Agents for
  Architecture Search},'' nov 2018.

\bibitem{Blakemore1970}
C.~Blakemore, R.~H.~S. Carpenter, and M.~A. Georgeson, ``{Lateral Inhibition
  between Orientation Detectors in the Human Visual System},'' {\em Nature},
  vol.~228, pp.~37--39, oct 1970.

\bibitem{Lambert1758}
J.~H. Lambert, ``{Observationes variae in mathesin puram},'' {\em Acta
  Helvetica}, vol.~3, no.~1, pp.~128--168, 1758.

\bibitem{Corless1996}
R.~M. Corless, G.~H. Gonnet, D.~E.~G. Hare, D.~J. Jeffrey, and D.~E. Knuth,
  ``{On the LambertW function},'' {\em Advances in Computational Mathematics},
  vol.~5, pp.~329--359, dec 1996.

\bibitem{Klimesch2012}
W.~Klimesch, ``{Alpha-band oscillations, attention, and controlled access to
  stored information},'' {\em Trends in Cognitive Sciences}, vol.~16,
  pp.~606--617, dec 2012.

\bibitem{Buzsaki2006}
G.~Buzsaki, {\em {Rhythms of the Brain}}.
\newblock Oxford University Press, 2006.

\bibitem{Kingma2014}
D.~P. Kingma and J.~Ba, ``{Adam: A Method for Stochastic Optimization},''
  pp.~1--15, dec 2014.

\bibitem{Glorot2010}
X.~Glorot and Y.~Bengio, ``{Understanding the Difficulty of Training Deep
  Feedforward Neural Networks},'' in {\em Proc. of the Int. Conf. on Artificial
  Intelligence and Statistics}, vol.~9, pp.~249--256, 2010.

\bibitem{Klimesch2007}
W.~Klimesch, P.~Sauseng, and S.~Hanslmayr, ``{EEG alpha oscillations: The
  inhibition–timing hypothesis},'' {\em Brain Research Reviews}, vol.~53,
  pp.~63--88, jan 2007.

\bibitem{Wang1996}
X.-J.~J. Wang and G.~Buzs{\'{a}}ki, ``{Gamma Oscillation by Synaptic Inhibition
  in a Hippocampal Interneuronal Network Model},'' {\em The Journal of
  neuroscience : the official journal of the Society for Neuroscience},
  vol.~16, pp.~6402--13, oct 1996.

\bibitem{Golovin2017}
D.~Golovin, B.~Solnik, S.~Moitra, G.~Kochanski, J.~Karro, and D.~Sculley,
  ``{Google Vizier},'' in {\em Proceedings of the 23rd ACM SIGKDD International
  Conference on Knowledge Discovery and Data Mining - KDD '17}, (New York, New
  York, USA), pp.~1487--1495, ACM Press, 2017.

\bibitem{Erhan2009}
D.~Erhan, Y.~Bengio, A.~Courville, and P.~Vincent, ``{Visualizing Higher-Layer
  Features of a Deep Network},'' tech. rep., D{\'{e}}partement d'informatique
  et de recherche op{\'{e}}rationnelle, 2009.

\bibitem{Mordvintsev2015}
A.~Mordvintsev, C.~Olah, and M.~Tyka, ``{Inceptionism: Going Deeper into Neural
  Networks},'' 2015.

\bibitem{Olah2017}
C.~Olah, A.~Mordvintsev, and L.~Schubert, ``{Feature visualization},'' {\em
  Distill}, vol.~2, no.~11, p.~e7, 2017.

\bibitem{ihmehimmeli2019}
``\url{https://github.com/google/ihmehimmeli}.''

\bibitem{OConnor2016}
P.~O'Connor and M.~Welling, ``{Deep Spiking Networks},'' pp.~1--16, feb 2016.

\bibitem{Tavanaei2019bpstdp}
A.~Tavanaei and A.~Maida, ``{BP-STDP: Approximating backpropagation using spike
  timing dependent plasticity},'' {\em Neurocomputing}, vol.~330, pp.~39--47,
  feb 2019.

\bibitem{Querlioz2013}
D.~Querlioz, O.~Bichler, P.~Dollfus, and C.~Gamrat, ``{Immunity to Device
  Variations in a Spiking Neural Network With Memristive Nanodevices},'' {\em
  IEEE Transactions on Nanotechnology}, vol.~12, pp.~288--295, may 2013.

\bibitem{Brader2007}
J.~M. Brader, W.~Senn, and S.~Fusi, ``{Learning Real-World Stimuli in a Neural
  Network with Spike-Driven Synaptic Dynamics},'' {\em Neural Computation},
  vol.~19, pp.~2881--2912, nov 2007.

\bibitem{Tavanaei2019deep}
A.~Tavanaei, M.~Ghodrati, S.~R. Kheradpisheh, T.~Masquelier, and A.~Maida,
  ``{Deep learning in spiking neural networks},'' {\em Neural Networks},
  vol.~111, pp.~47--63, mar 2019.

\bibitem{Ward2003}
L.~M. Ward, ``{Synchronous neural oscillations and cognitive processes},'' {\em
  Trends in Cognitive Sciences}, vol.~7, pp.~553--559, dec 2003.

\bibitem{Shanahan2008}
M.~Shanahan, ``{A spiking neuron model of cortical broadcast and
  competition},'' {\em Consciousness and Cognition}, vol.~17, pp.~288--303, mar
  2008.

\bibitem{Gamez2009}
D.~Gamez, ``{Information integration based predictions about the conscious
  states of a spiking neural network},'' {\em Consciousness and Cognition},
  vol.~19, pp.~294--310, mar 2010.

\bibitem{Davies2018}
M.~Davies, N.~Srinivasa, T.-H. Lin, G.~Chinya, Y.~Cao, S.~H. Choday, G.~Dimou,
  P.~Joshi, N.~Imam, S.~Jain, Y.~Liao, C.-K. Lin, A.~Lines, R.~Liu,
  D.~Mathaikutty, S.~McCoy, A.~Paul, J.~Tse, G.~Venkataramanan, Y.-H. Weng,
  A.~Wild, Y.~Yang, and H.~Wang, ``{Loihi: A Neuromorphic Manycore Processor
  with On-Chip Learning},'' {\em IEEE Micro}, vol.~38, pp.~82--99, jan 2018.

\bibitem{Pfeiffer2018}
M.~Pfeiffer and T.~Pfeil, ``{Deep Learning With Spiking Neurons: Opportunities
  and Challenges},'' {\em Frontiers in Neuroscience}, vol.~12, oct 2018.

\bibitem{Berry1997}
M.~J. Berry, D.~K. Warland, and M.~Meister, ``{The structure and precision of
  retinal spike trains},'' {\em Proceedings of the National Academy of
  Sciences}, vol.~94, pp.~5411--5416, may 1997.

\bibitem{Palva2016}
S.~Palva and J.~M. Palva, ``{The Role of Local and Large-Scale Neuronal
  Synchronization in Human Cognition},'' in {\em Multimodal Oscillation-based
  Connectivity Theory}, pp.~51--67, Cham: Springer International Publishing,
  2016.

\bibitem{Diesmann1999}
M.~Diesmann, M.-O. Gewaltig, and A.~Aertsen, ``{Stable propagation of
  synchronous spiking in cortical neural networks},'' {\em Nature}, vol.~402,
  pp.~529--533, dec 1999.

\bibitem{Lestienne2001}
R.~Lestienne, ``{Spike timing, synchronization and information processing on
  the sensory side of the central nervous system},'' {\em Progress in
  Neurobiology}, vol.~65, pp.~545--591, dec 2001.

\bibitem{Engel1992}
A.~K. Engel, P.~K{\"{o}}nig, A.~K. Kreiter, T.~B. Schillen, and W.~Singer,
  ``{Temporal coding in the visual cortex: new vistas on integration in the
  nervous system.},'' {\em Trends in neurosciences}, vol.~15, pp.~218--26, jun
  1992.

\bibitem{Gray1999}
C.~M. Gray, ``{Visual Feature Integration : Still Alive and Well},'' {\em
  Neuron}, vol.~24, pp.~31--47, 1999.

\bibitem{Ahissar1998}
E.~Ahissar, ``{Temporal-Code to Rate-Code Conversion by Neuronal Phase-Locked
  Loops},'' {\em Neural Computation}, vol.~10, pp.~597--650, apr 1998.

\bibitem{Mehta2002}
M.~R. Mehta, A.~K. Lee, and M.~A. Wilson, ``{Role of experience and
  oscillations in transforming a rate code into a temporal code},'' {\em
  Nature}, vol.~417, pp.~741--746, jun 2002.

\bibitem{Crick1989}
F.~Crick, ``{The recent excitement about neural networks},'' {\em Nature},
  vol.~337, pp.~129--132, jan 1989.

\bibitem{Hassabis2017}
D.~Hassabis, D.~Kumaran, C.~Summerfield, and M.~Botvinick,
  ``{Neuroscience-Inspired Artificial Intelligence},'' {\em Neuron}, vol.~95,
  pp.~245--258, jul 2017.

\bibitem{Liao2015}
Q.~Liao, J.~Z. Leibo, and T.~Poggio, ``{How Important is Weight Symmetry in
  Backpropagation?},'' pp.~1837--1844, oct 2015.

\bibitem{Bengio2014}
Y.~Bengio, ``{How Auto-Encoders Could Provide Credit Assignment in Deep
  Networks via Target Propagation},'' pp.~1--34, jul 2014.

\bibitem{Lillicrap2016}
T.~P. Lillicrap, D.~Cownden, D.~B. Tweed, and C.~J. Akerman, ``{Random synaptic
  feedback weights support error backpropagation for deep learning},'' {\em
  Nature Communications}, vol.~7, p.~13276, dec 2016.

\bibitem{Whittington2017}
J.~C.~R. Whittington and R.~Bogacz, ``{An Approximation of the Error
  Backpropagation Algorithm in a Predictive Coding Network with Local Hebbian
  Synaptic Plasticity},'' {\em Neural Computation}, vol.~29, pp.~1229--1262,
  may 2017.

\bibitem{Song2000}
S.~Song, K.~D. Miller, and L.~F. Abbott, ``{Competitive Hebbian learning
  through spike-timing-dependent synaptic plasticity},'' {\em Nature
  Neuroscience}, vol.~3, pp.~919--926, sep 2000.

\bibitem{Bengio2015}
Y.~Bengio, D.-H. Lee, J.~Bornschein, T.~Mesnard, and Z.~Lin, ``{Towards
  Biologically Plausible Deep Learning},'' feb 2015.

\bibitem{Newman2003}
E.~A. Newman, ``{New roles for astrocytes: Regulation of synaptic
  transmission},'' {\em Trends in Neurosciences}, vol.~26, no.~10,
  pp.~536--542, 2003.

\bibitem{Eroglu2010}
C.~Eroglu and B.~A. Barres, ``{Regulation of synaptic connectivity by glia},''
  {\em Nature}, vol.~468, pp.~223--231, nov 2010.

\end{thebibliography}
\bibliographystyle{ieeetr}

\appendix
\section{Appendix}
\label{apx:ua}

Here we prove that the proposed spiking network model is an universal
approximator for sufficiently well-behaved functions.

Recall that the maximum voltage of a spike of positive weight $w$ is obtained
$\frac{1}{\tau}$ units of time after the spike itself, reaching a value of
$\frac{w}{\tau e}$.

\begin{lemma}
For every~$\epsilon>0$, here is a spiking network with $n$ inputs, each of them
  constrained to the $[0, 1]$ range, and a single output that produces a spike
  in the interval $(t, t+\epsilon)$, with $t \ge 2+\frac{2}{\tau}$ if and only
  if each of the inputs belongs to a specified interval. Otherwise, no spike is
  produced.  Moreover, it is possible to build one such network by using $2n+4$
  neurons.\label{lem:box}

\end{lemma}
\begin{proof}
We will first prove:
\begin{lemma}
  We can use one neuron and three auxiliary pulse inputs to conditionally
  produce a spike in the time interval $(t_{out}, t_{out} + \epsilon)$ if and
  only if the input spike happens at a time $t_i$ that is smaller than (or
  bigger than) a given constant $t_0 < t_{out}$, provided that $t_{out} > 2 +
  \frac{1}{\tau}$.  \label{lem:thres}
\end{lemma}

\begin{proof}

Let us consider the function:

$\Delta_x(t) = w(t e^{-\tau t} - (t-T)e^{-\tau t + \tau T}) = w e^{-\tau t} (t
  - (t - T) e^{\tau T})$

This describes the potential of a neuron that receives one spike at time $0$
  with positive weight $w$, and one at time $T \in [-1, 1]$ with negative
  weight $-w$; note that this formula is valid for $t \ge \max\{0, T\}$.  The
  sign of this function is determined fully by the term $t - (t-T)e^{\tau T}$.
  Note that $t - (t-T)e^{\tau T} > 0 \Leftrightarrow t (e^{\tau T}-1) < T
  e^{\tau T}$.

If $T < 0$, then $e^{\tau T} < 1$, thus $t (e^{\tau T}-1) < T e^{\tau T}
  \Leftrightarrow t > -T \frac{e^{\tau T}}{1-e^{\tau T} }$. As $-T
  \frac{e^{\tau T}}{1-e^{\tau T} } < \frac{1}{\tau}$ for $T \in [-1, 0)$, it
  follows that if $t>\frac{1}{\tau}$ then $\Delta_T(t)>0$.

If $T > 0$, we have $t (e^{\tau T}-1) > T e^{\tau T} \Leftrightarrow t > T
  \frac{e^{\tau T}}{e^{\tau T} -1}$. As $T \frac{e^{\tau T}}{e^{\tau T} -1} < 1
  + \frac{1}{\tau}$, it follows that if $t > 1+\frac{1}{\tau}$ then
  $\Delta_T(t)<0$.

Thus, we can consider this setup:
\begin{itemize}
    \item an input spike at time $t_i \in [0, 1]$ with positive weight $w$,
    \item a fixed spike at time $t_0 \in [0, 1]$ with negative weight $w <
      \theta \tau e$,
    \item a fixed spike at time $t_{out} > 2 + \frac{1}{\tau}$ with weight $v \ge
      \theta \tau e$,
    \item a fixed spike at the time when the spike starting at $t_{out}$ would
      reach the firing threshold if no other spikes were present, with a big
      enough negative weight to ensure that that spike reaches but does not
      cross the threshold by itself.
\end{itemize}
If $t_i\le t_0$, then at time $t_{out}$ the potential of the neuron is
  positive (as proven above). Thus, the fixed spike at $t_0$ will cross the
  firing threshold and produce an output spike after a small delay (which can
  be made smaller than $\epsilon$ by suitably increasing $v$). On the other
  hand, if $t_i>t_0$, the potential will be negative, and no output spike is
  produced.

In a similar way, we can have a configuration of neurons and weights that will
  produce a spike in the interval $(t_{out}, t_{out}+\epsilon)$ if and only if
  $t_i \ge t_0$.
\end{proof}

By connecting $2n$ of the configurations shown in Lemma~\ref{lem:thres} to a
  single neuron, having connections with weight $\frac{\theta\tau e}{2n-1} -
  \delta$ for a fixed $\delta>0$ that depends on the desired $\epsilon$, we
  obtain an output spike in the interval $(t_{out} + \frac{1}{\tau}, t_{out} +
  \frac{1}{\tau} + \epsilon$) if and only if the input belongs to a specified
  product of intervals. This requires the output spikes of
  Lemma~\ref{lem:thres} to belong to an interval that is small enough to ensure
  that any configuration of input spikes will still reach a potential of
  $\theta$. This is always possible if the chosen $\epsilon$ in
  Lemma~\ref{lem:thres} is small enough.

  Note that
  in the configuration from Lemma~\ref{lem:thres} all the fixed input spikes
  can be shared, except for the one defining the threshold. Thus, this setup
  requires at most $2n+3+1$ neurons.
\end{proof}

\begin{theorem}
Let $f$ be a continuous function from $[0, 1]^n$ to $(2+\frac{2}{\tau},
  +\infty)$. Then $\forall \epsilon > 0$, there is a spiking network with $2$
  hidden layers and  decay constant $\tau$ that computes a function $g$
  satisfying $|f(x) - g(x)| < \epsilon\ \forall x \in [0, 1]^n$.

If $f$ is also Lipschitz with constant $K$, then it is possible to realize such
  a network using at most $(3K\sqrt{n}\epsilon^{-1})^n(4n+3)+1$ neurons.
\end{theorem}

\begin{proof}
Since $f$ is continuous on a compact set, it is uniformly continuous. Thus, for
  any $\epsilon>0$, there is a $\delta$ such that, in each subdivision of the
  domain in $n$-dimensional boxes with extension along each coordinate axis
  being no more than $\delta$, the difference between the maximum and the
  minimum of $f$ is at most $\frac{\epsilon}{3}$.  Moreover, if the function is
  Lipschitz with constant $K$, we can choose $\frac{\epsilon}{3}/(K\sqrt n)$.

By Lemma~\ref{lem:box}, for each such box $B$ there is a network that produces
  a spike in the time interval $(\min_Bf, \min_Bf + \frac{\epsilon}{3})$ if and
  only if the input belongs to the box. Connecting the outputs of those
  networks to a single neuron with sufficiently high weights, we can ensure
  that it will produce a spike with a delay of at most $\frac{\epsilon}{3}$
  after receiving one as an input.

Thus, as a whole the network will produce a spike for input $x$ with a time
  that is at most $\epsilon$ away from the value $f(x)$, proving the thesis.
\end{proof}

Note that the requirement for the function domain to be greater than some
prefixed values is necessary: any network that uses temporal coding, takes two
inputs $x, y$, and produces as an output $\frac{x+y}{2}$ would violate
causality.

\end{document}